\documentclass[11pt]{article}

\usepackage[preprint]{acl}
\usepackage{arydshln}
\usepackage{caption} 
\usepackage{wrapfig}
\usepackage{dblfloatfix}
\usepackage{tcolorbox}
\usepackage{subcaption}
\usepackage{xspace}  
\usepackage{dblfloatfix} 
\usepackage{hyperref}       
\usepackage{url}            
\usepackage{booktabs}       
\usepackage{amsfonts}       
\usepackage{nicefrac}       
\usepackage{enumitem}
\usepackage{amsmath}
\usepackage{multirow}
\usepackage[table]{xcolor}

\usepackage{times}
\usepackage{latexsym}

\usepackage[T1]{fontenc}

\usepackage[utf8]{inputenc}

\usepackage{microtype}

\usepackage{inconsolata}

\usepackage{graphicx}

\title{\ours: Benchmarking Foundation Models on Reasoning and Planning with Puzzles}

\addtocontents{toc}{\protect\setcounter{tocdepth}{-1}}
\newcommand{\ours}{\textsc{PuzzlePlex}\xspace}

\author{Yitao Long$^{1}$ \quad Yuru Jiang$^{2}$ \quad Hongjun Liu$^{1}$ \quad Yilun Zhao$^{3}$ \quad Jingchen Sun$^{4}$ \\ \bf{Yiqiu Shen$^{1,5}$  \quad Chen Zhao$^{1}$ \quad Arman Cohan $^{3}$ \quad Dennis Shasha$^{1}$}  
\vspace{4pt}\\
$^1$New York University \quad $^2$Zhejiang University \quad $^3$Yale University \\ \quad $^4$University at Buffalo, SUNY \quad $^5$NYU Grossman School of Medicine \vspace{5pt}\\
\github~~~\href{https://github.com/xxx/xxx}{\path{https://github.com/yitaoLong/PuzzlePlex}}\\
}

\usepackage{xcolor}  
\usepackage{amssymb} 
\usepackage{pifont}  
\newcommand{\cmark}{\textcolor{green!50!black}{\checkmark}}   
\newcommand{\xmark}{\textcolor{red}{\ding{55}}}   
\newcommand{\github}{\raisebox{-1.5pt}{\includegraphics[height=1.05em]{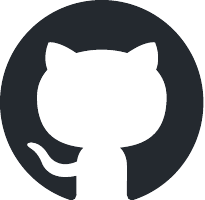}}\xspace}
\begin{document}
\maketitle

\begin{abstract}
This work investigates the reasoning and planning capabilities of foundation models and their scalability in complex, dynamic environments. We introduce \ours, a benchmark designed to assess these capabilities through a diverse set of puzzles. \ours consists of 15 types of puzzles, including deterministic and stochastic games of varying difficulty, as well as single-player and two-player scenarios. The \ours framework provides a comprehensive environment for each game, and supports extensibility to generate more challenging instances as foundation models evolve. Additionally, we implement customized game-playing strategies for comparison.  
Building on this benchmark, we develop fine-grained metrics to measure performance and conduct an in-depth analysis of frontier foundation models across two settings: \emph{instruction-based} and \emph{code-based}. Furthermore, we systematically investigate their scaling limits. Our findings show that reasoning models outperform others in instruction-based settings, while code-based execution presents greater challenges but offers a scalable and efficient alternative. \ours enables targeted evaluation and guides future improvements in reasoning, planning, and generalization for foundation models. 
\end{abstract}

\section{Introduction} \label{sec:intro}
The rapid progress of foundation models has led to remarkable improvements across a broad spectrum of natural language processing tasks. Recently, the emergence of reasoning models such as OpenAI o-series models \cite{openai2024openaio1card} and DeepSeek-R1 \cite{deepseekai2025deepseekr1incentivizingreasoningcapability} have demonstrated
remarkable advances in complex reasoning tasks through test-time compute scaling. 
These breakthroughs naturally prompt a deeper question: \emph{How far can modern models push genuine problem-solving ability, especially in scenarios that demand sustained, structured reasoning?}

\begin{figure}[t]
    \centering
    \includegraphics[width = 1.0\linewidth]{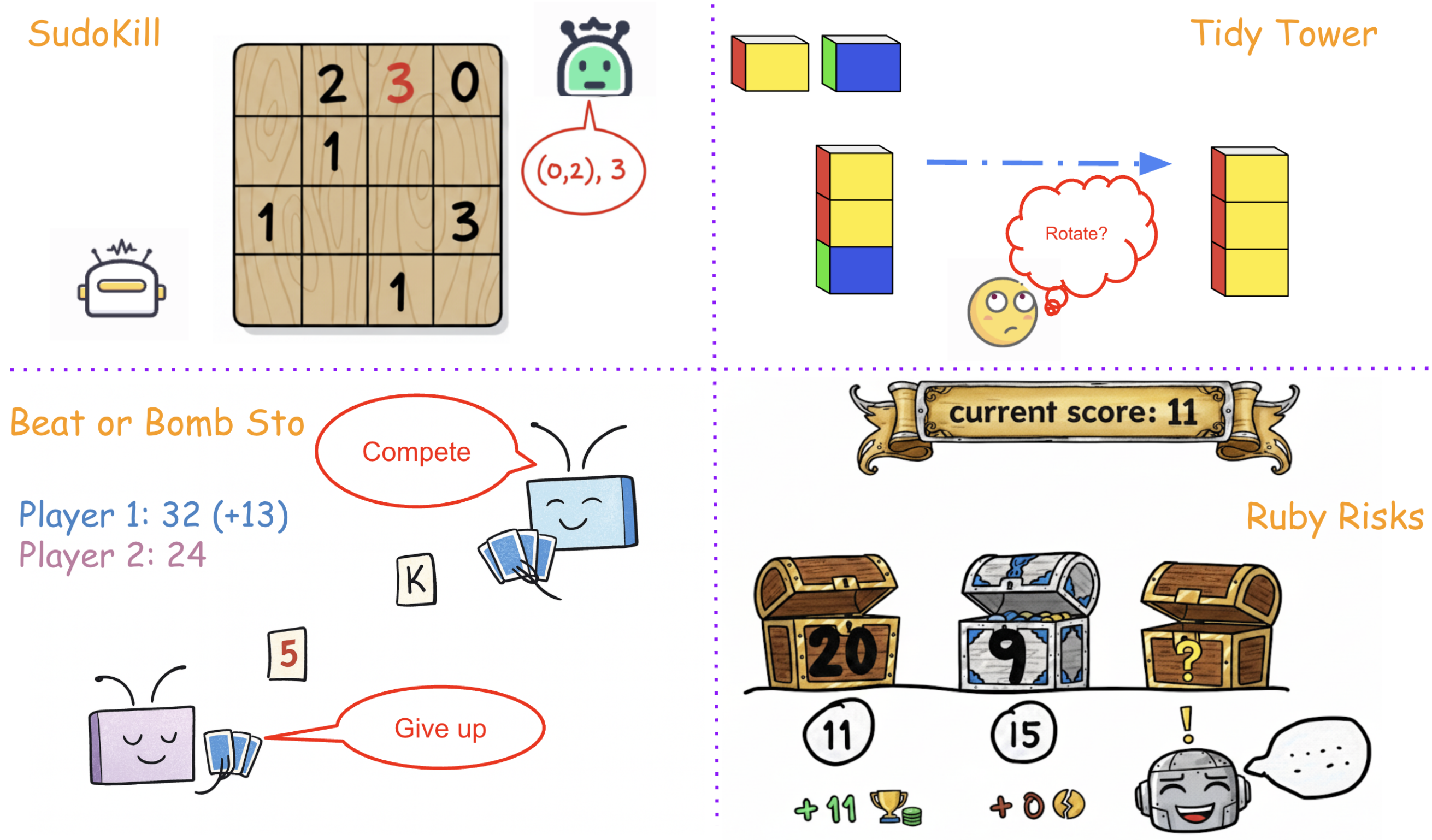}
    \caption{Overview of four puzzles: \textbf{SudoKill} (two-player deterministic), \textbf{Tidy Tower} (single-player deterministic), \textbf{Beat or Bomb Sto} (two-player stochastic), and \textbf{Ruby Risks} (single-player stochastic).}
    \label{fig:puz_plex_example}
    \vspace{-1em}
\end{figure}

To explore this, we turn to puzzle solving—a domain that inherently blends logical, numerical, and spatial reasoning with long-horizon planning and strategic adaptation. Many puzzles unfold over multiple interactive steps, involving competition or dynamic environments. This makes them ideal for evaluating a model’s ability to reason under evolving constraints, adapt to new strategies, and maintain coherence across extended interactions.

\begin{table*}[t]
\centering
\vspace{5pt}
\addtolength{\tabcolsep}{-0.35em}
\resizebox{\textwidth}{!}{%
\begin{tabular}{@{}lcccccccc@{}ccc}
\toprule
\multirow{2}{*}{\textbf{Benchmark}} & \multicolumn{2}{c}{\textbf{Game Scenario}} & \multicolumn{2}{c}{\textbf{Reward Predictability}} & \textbf{\# Multi-} & \multicolumn{2}{c}{\textbf{Data Type}} & \textbf{Varying} && \multicolumn{2}{c}{\textbf{Evaluation}} \\
\cmidrule(lr){2-3} \cmidrule(lr){4-5} \cmidrule(lr){7-8} \cmidrule(lr){11-12}
 & {Single-} & Two-player & Deterministic & Stochastic & \textbf{Turn} & Text & Text-Image & \textbf{Difficulty} & &Inst.  & Code \\
\midrule
\textsc{PuzzleBench} \cite{mittal2025fcorebenchlargelanguagemodels} & \cmark & \xmark & \cmark & \xmark & \xmark & \cmark & \xmark & \xmark &&\xmark &\cmark \\
\textsc{PUZZLES} \cite{NEURIPS2024_e5d1eaad} & \cmark & \xmark & \cmark &\xmark &\cmark & \xmark &\cmark &\cmark &&\xmark &\cmark  \\
\textsc{LogicGame} \cite{gui2024logicgame}     & \cmark & \xmark & \cmark & \xmark & \xmark & \cmark & \xmark & \xmark &&\cmark &\xmark \\
\textsc{BoardgameQA} \cite{kazemi2023boardgameqa} & \cmark & \xmark & \cmark & \cmark & \xmark & \cmark & \xmark & \xmark &&\cmark &\xmark \\
\textsc{P3} \cite{schuster2021programming}       & \cmark & \xmark & \cmark & \xmark & \xmark & \cmark & \xmark & \cmark &&\xmark &\cmark \\
\textsc{PUZZLEQA} \cite{zhao2023solving} & \cmark & \xmark & \cmark & \xmark & \xmark & \xmark & \xmark & \xmark &&\cmark &\xmark\\
\textsc{EnigmaEval} \cite{wang2025enigmaevalbenchmarklongmultimodal} &\cmark &\xmark &\cmark &\xmark &\cmark &\cmark &\cmark &\cmark &&\cmark &\xmark\\
\textsc{VGRP-Bench} \cite{ren2025vgrpbenchvisualgridreasoning} & \cmark & \xmark & \cmark & \cmark &\cmark &\cmark &\cmark &\cmark &&\cmark &\xmark\\
\textsc{SmartPlay} \cite{wu2023smartplay}      & \cmark & \cmark & \cmark & \cmark & \cmark & \cmark & \xmark & \xmark &&\cmark &\xmark\\
\midrule
\textsc{\ours} (ours)        & \cmark & \cmark & \cmark & \cmark & \cmark & \cmark & \cmark & \cmark &&\cmark &\cmark\\
\bottomrule
\end{tabular}%
}
\caption{Comparison between \ours and existing puzzle benchmarks. A single-turn game ends after one move by one or more players.}
\label{tab:puzzleplex_comparison}
\vspace{-1.5em}
\end{table*}

To this end, we introduce \ours, a benchmark designed to evaluate foundation models' reasoning and planning capabilities.
Unlike prior benchmarks \cite{zhang2025puzzlebenchfullydynamicevaluation, wu2024smartplaybenchmarkllmsintelligent} that reuse common puzzles—many potentially seen during pretraining—\ours features 15 novel, curated puzzles spanning both \textbf{text-only} and \textbf{text-image} formats. As shown in \autoref{fig:puz_plex_example}, the puzzles cover \textbf{single-player} and \textbf{two-player} settings and include both \textbf{deterministic} and \textbf{stochastic} environments. Each puzzle supports multiple difficulty levels and extensible generation, enabling adaptive evaluation as models improve. Their long-horizon, dynamic nature provides a compact yet demanding testbed for assessing reasoning depth, planning, and strategic coherence—areas underexplored in prior short-context benchmarks \cite{gui2024logicgamebenchmarkingrulebasedreasoning}.

We further design hand-crafted strategies for comparison and evaluate models under two complementary paradigms: \textbf{instruction-based} and \textbf{code-based}. In the former, models act as agents interacting via natural language; in the latter, they generate executable code that solves the puzzle. Together, these paradigms reveal both interactive reasoning and programmatic abstraction capabilities.

Empirical results show that reasoning models outperform non-reasoning ones in instruction-based settings, leveraging test-time scaling and extended deliberation. However, performance drops in code-based evaluation due to challenges in program synthesis, though sampling-based methods help narrow the gap. Open-source models increasingly rival proprietary systems, and visual or legality-aware prompting further boosts results. However, models still struggle with multi-hop reasoning in some puzzles, suggesting limitations in their ability to maintain coherent reasoning over extended contexts.

In summary, our contributions are: 
\begin{itemize}[leftmargin=*]
\itemsep0em 
\item \ours, the first benchmark to jointly evaluate reasoning in both interactive and executable settings across diverse puzzle types.
\item Framework that supports textual and visual puzzles with deterministic and stochastic dynamics.
\item Hand-crafted baselines and fine-grained metrics enabling rigorous evaluation and comparison of systems and reasoning strategies.
\item Comprehensive empirical analysis across leading models, comparing the performance of different reasoning strategies, the scaling behavior of different systems, and the systems' failure modes.
\end{itemize}

\section{Related Work}
\subsection{Puzzles and Relevant Benchmarks}
Puzzles can be broadly divided into \textbf{rule-based} and \textbf{rule-less} types.
Rule-based puzzles, such as \textsc{Sudoku} \cite{noever2021puzzle}, \textsc{Crosswords} \cite{sadallah2024llms}, and \textsc{Chess} \cite{feng2023chessgpt}, have explicit rules, defined goals, and structured state transitions, requiring strategic and logical reasoning.
Rule-less puzzles, including Riddles \cite{lin2021riddlesense,bisk2020piqa}, lack explicit action spaces or clear objectives.
\ours focuses on \textbf{rule-based} puzzles to enable objective evaluation of reasoning in competitive, dynamic settings. We exclude knowledge-heavy puzzles (e.g., \textsc{Guess My City} \cite{abdulhai2023lmrl}) that depend on external knowledge \cite{schuster2021programming,lin2021riddlesense,todd2024missed}, since modern models already trained extensively on factual corpora and outperform humans on such tasks.

\autoref{tab:puzzleplex_comparison} compares \ours with recent puzzle benchmarks. Most existing benchmarks focus on \textbf{single-player}, short-horizon puzzles \cite{mittal2025fcorebenchlargelanguagemodels,gui2024logicgame,zhao2023solving}, while multi-turn, competitive two-player settings are rarely explored \cite{wu2023smartplay,liu2023agentbenchevaluatingllmsagents}. Few benchmarks incorporate \textbf{stochastic environments} for reasoning under uncertainty, or \textbf{multimodal puzzles} that require joint text-image understanding \cite{NEURIPS2024_e5d1eaad,wang2025enigmaevalbenchmarklongmultimodal,ren2025vgrpbenchvisualgridreasoning}.

\begin{figure}[t]
    \centering
    \includegraphics[width = 1.0\linewidth]{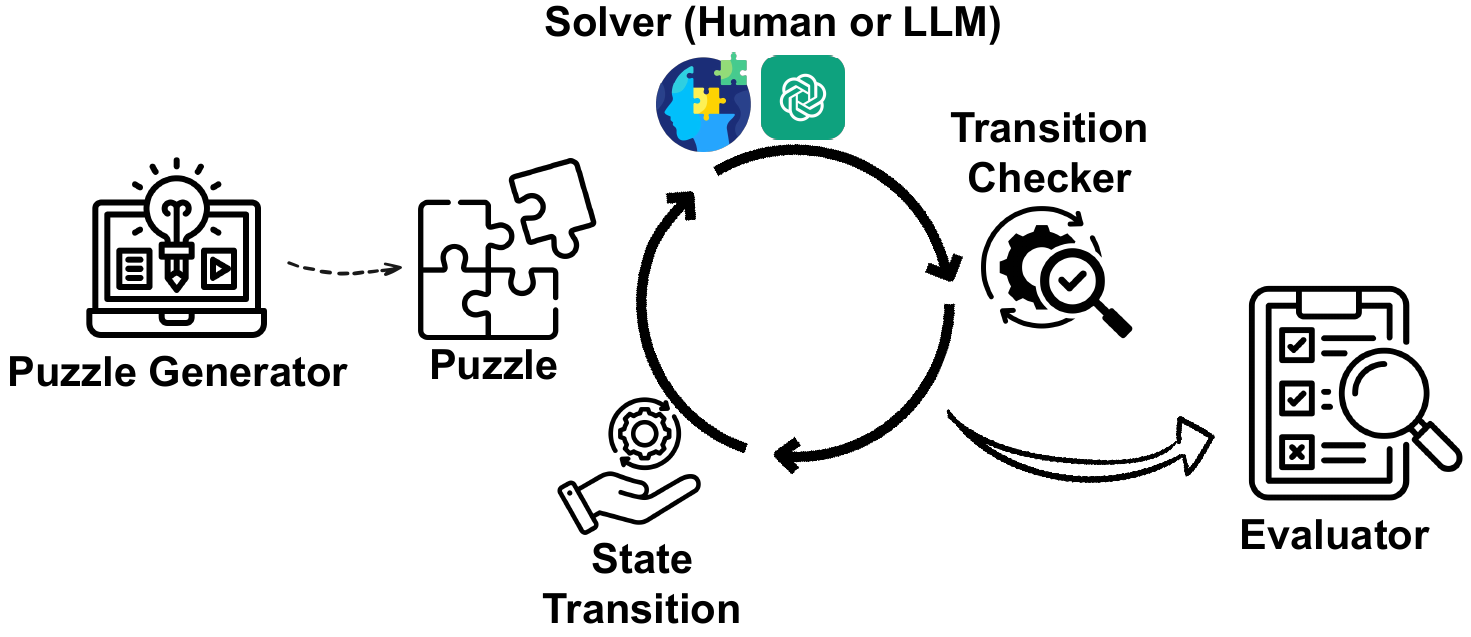}
    \caption{Overview of the developed pipeline framework. \textbf{Puzzle Generator} creates puzzle instances from templates based on the puzzle name, difficulty level, and selected competing models. The \textbf{Solver} then generates a response after receiving the puzzle instance. This response is passed to the \textbf{Transition Checker}, which verifies the legality of the operation output by the \textbf{Solver} and checks the game status. If the game ends, the \textbf{Evaluator} calculates and outputs the score. Otherwise, \textbf{State Transition} updates the state and passes the updated information back to the \textbf{Solver}.    }
    \label{fig:puz_plex_pipeline}
    \vspace{-1.5em}
\end{figure}

\subsection{Evolution of Puzzle Solving Techniques}
A variety of methods have been developed for solving rule-based puzzles.
Classical approaches rely on algorithmic techniques such as dynamic programming \cite{SMITH20071299}, alpha–beta pruning \cite{KORF1990189}, and heuristic search \cite{Lewis2007MetaheuristicsCS}.
For single-player puzzles, neurosymbolic methods \cite{ahmed2023semanticstrengtheningneurosymboliclearning,Murali_2022} are effective due to their combinatorial nature, often reducible to SAT or SMT formulations \cite{Bright_2020,hofler2014smt}.

With deep learning advances, reinforcement learning (RL) has become the dominant paradigm \cite{dos2022heuristics,huang2024pokergptendtoendlightweightsolver}, though combinatorial explosion still necessitates heuristics \cite{silver2016mastering}.
Early model-based approaches fine-tuned models like GPT-2 \cite{radford2019language} and FLAN-PaLM \cite{chung2024scaling} for puzzles such as Sudoku \cite{noever2021puzzle} and BoardgameQA \cite{kazemi2023boardgameqa}.
Stronger foundation models \cite{achiam2023gpt,anthropic2024claude} now solve puzzles through few-shot in-context learning and multi-run feedback.
Among prompting methods, Chain-of-Thought (CoT) \cite{wei2023chainofthoughtpromptingelicitsreasoning} consistently outperforms direct prompting, while extensions like Self-Refine \cite{madaan2023selfrefineiterativerefinementselffeedback}, Tree-of-Thought (ToT) \cite{yao2023tree}, and Everything-of-Thoughts \cite{ding2023everything} further enhance reasoning for deterministic puzzles.

\begin{table*}[t]
\centering
\small
\small
\begin{tabular}{lcccccc}
\toprule
\multirow{2}{*}{\textbf{Model}} &
\multicolumn{2}{c}{\textbf{Single-Player Det.}} &
\multicolumn{2}{c}{\textbf{Two-Player Det.}} &
\multirow{2}{*}{\textbf{Score}} \\
\cmidrule(lr){2-3}  \cmidrule(lr){4-5}
 & Easy & Normal & Easy & Normal \\
\midrule
Custom & 0.89 \tiny{$\pm$0.47} & 0.83 \tiny{$\pm$0.74} & 0.59 \tiny{$\pm$0.36} & 0.60 \tiny{$\pm$0.34} & 0.70 \tiny{$\pm$0.15} \\
\midrule
Deepseek-R1 & \cellcolor{red!35}{0.64 \tiny{$\pm$1.38}} & \cellcolor{red!35}{0.48 \tiny{$\pm$1.04}} & 0.66 \tiny{$\pm$0.12} & \cellcolor{red!5}{0.66 \tiny{$\pm$0.14}} & \cellcolor{red!35}{0.62 \tiny{$\pm$0.15}} \\
o4-mini & \cellcolor{red!5}{0.44 \tiny{$\pm$1.16}} & \cellcolor{red!20}{0.44 \tiny{$\pm$1.14}} &\cellcolor{red!3}{ 0.67 \tiny{$\pm$0.15}} & \cellcolor{red!35}{0.68 \tiny{$\pm$0.12}} & \cellcolor{red!20}{0.59 \tiny{$\pm$0.15}} \\
Gemini-2.5-pro & \cellcolor{red!5}{0.44 \tiny{$\pm$1.13}} & \cellcolor{red!20}{0.44 \tiny{$\pm$1.02}} & \cellcolor{red!20}{0.68 \tiny{$\pm$0.14}} & \cellcolor{red!20}{0.67 \tiny{$\pm$0.13}} & \cellcolor{red!5}{0.58 \tiny{$\pm$0.14}} \\
QwQ-32B & \cellcolor{red!20}{0.54 \tiny{$\pm$1.15}} & 0.25 \tiny{$\pm$0.58} & \cellcolor{red!35}{0.69 \tiny{$\pm$0.10}} & \cellcolor{red!35}{0.68 \tiny{$\pm$0.11}} & \cellcolor{red!5}{0.58 \tiny{$\pm$0.14}} \\
grok-3-mini & 0.17 \tiny{$\pm$0.52} & 0.22 \tiny{$\pm$0.51} & \cellcolor{red!5}{0.67 \tiny{$\pm$0.20}} &  \cellcolor{red!20}{0.67 \tiny{$\pm$0.20}} & 0.49 \tiny{$\pm$0.15} \\
Deepseek-V3 & 0.34 \tiny{$\pm$0.78} & 0.24 \tiny{$\pm$0.54} & 0.52 \tiny{$\pm$0.13} & 0.50 \tiny{$\pm$0.14} & 0.43 \tiny{$\pm$0.10} \\
GPT-4.1 & 0.40 \tiny{$\pm$0.98} & \cellcolor{red!5}{0.35 \tiny{$\pm$0.94}} & 0.44 \tiny{$\pm$0.13} & 0.45 \tiny{$\pm$0.09} & 0.42 \tiny{$\pm$0.11} \\
Qwen-2.5-VL-72B & 0.24 \tiny{$\pm$0.62} & 0.24 \tiny{$\pm$0.37} & 0.24 \tiny{$\pm$0.18} & 0.28 \tiny{$\pm$0.24} & 0.25 \tiny{$\pm$0.09} \\
Llama-3.3-70B & 0.15 \tiny{$\pm$0.37} & 0.12 \tiny{$\pm$0.39} & 0.31 \tiny{$\pm$0.12} & 0.31 \tiny{$\pm$0.12} & 0.25 \tiny{$\pm$0.07} \\
Gemma-3-27B & 0.13 \tiny{$\pm$0.38} & 0.12 \tiny{$\pm$0.39} & 0.23 \tiny{$\pm$0.12} & 0.24 \tiny{$\pm$0.10} & 0.19 \tiny{$\pm$0.06} \\
Phi-4-multimodal & 0.05 \tiny{$\pm$0.20} & 0.03 \tiny{$\pm$0.14} & 0.17 \tiny{$\pm$0.12} & 0.17 \tiny{$\pm$0.05} & 0.12 \tiny{$\pm$0.05} \\
\bottomrule
\end{tabular}
\caption{Instruction-based normalized scores (mean $\pm$ 95\% CI) of models on single-player and two-player deterministic puzzles, separated by difficulty. }
\label{tab:det_puzzle_scores}
\end{table*}

In this work, we adopt CoT-style prompting to evaluate systematic reasoning, planning, and decision-making.
We further introduce a \textbf{code-based execution} setting, where models generate and execute code to interact directly with puzzle environments—linking reasoning with concrete actions and improving solution correctness and generalization.

\section{\ours}
We first introduce the \ours framework in which puzzle templates can be instantiated, moves recorded, state information shared, and states evaluated. We next describe the puzzles included in this benchmark, the implementation of customized strategies, and the evaluation methods.

\subsection{Puzzle Generation Framework}
\label{framework}
 
\ours has the following main components, as presented in \autoref{fig:puz_plex_pipeline}.

\paragraph{Instance Generation.} For each puzzle $p$, we distinguish between a possibly parametrized puzzle template $template(p)$ (e.g., SudoKill on a 9 $\times$ 9 grid, template(SudoKill(9,9)), and an instance $instance(p)$ (e.g., a particular instance of SudoKill on a 9 $\times$ 9 grid, instance(Suduoku(9,9)). A generator function $G_p$ maps templates to instances. The generated instance is also the initial state $S_0$ of the game. That is, $instance(p)=S_0$. 
The generator for each puzzle  will  create instances using randomness, and it will adjust the difficulty level by varying the size of the puzzle.

\paragraph{State Transition.} After receiving a move $M$ generated by a player (human or computer), the state transition module maps a state $S_n$ to a new state $S_{n+1}$ while incorporating feedback $F_n$. The feedback $F_n$ indicates the legality of the move, whether the game has terminated, and provides new position information. This process is represented as $M: ~ S_n \rightarrow (S_{n+1}, F_n)$.

\paragraph{Evaluation.} Once the puzzle-solving process terminates, an \textbf{Evaluator} $E_p$ is applied to the sequence of states $S_0, S_1, \ldots, S_n$ to determine the raw score(s), represented as $rs_p = E_p(S_0, S_1, \ldots, S_n)$. The scale of the raw scores varies depending on the resolution type of each puzzle. To ensure comparability, we normalize these scores to obtain final scores ranging from 0 to 1 (\S~\ref{evaluation}).

To better keep track of state transitions and model reasoning steps, we implemented a Web UI called \textbf{Simulator} for visual observation. An example of this interface is shown in the \S ~\ref{simulator}.

\subsection{\ours Benchmark Construction}

All puzzles in \ours are either derived from a column in Communications of the ACM \footnote{https://cacm.acm.org/section/opinion/} or manually curated by the authors. While foundation models may have been exposed to textual descriptions of these puzzles, there are no publicly available strategies for solving them, thereby minimizing the risk of data contamination during gameplay. Additionally, we have simplified the rules of several puzzles to reduce the barrier to entry, enabling most users to engage with them immediately after learning the rules and objectives.

Our 15 puzzles are categorized into four types: \textbf{single-player deterministic}, \textbf{single-player stochastic}, \textbf{two-player deterministic}, and \textbf{two-player stochastic}. Text-based puzzles span all four types, whereas text-image puzzles are limited to the two-player deterministic type.
The distinction between deterministic and stochastic games lies in the predictability of operation outcomes. In deterministic games, the result of a decision is fixed, regardless of how many times it is taken. In contrast, stochastic games produce probabilistic outcomes, where repeated execution of the same operation in the same state may lead to different results. Detailed information about the puzzles is provided in \S\ref{dataset}, and individual puzzle descriptions are included in \S\ref{description}.

\subsection{Customized Strategies}
We implemented customized strategies for each puzzle, which can be categorized as follows:
\begin{itemize}[leftmargin=*]
\itemsep0em 
\item \textbf{Brute-force Algorithm}: This method is employed when the problem size allows for an exhaustive search within our specified time constraints.

\item \textbf{Search Algorithms}: We employ a variety of search techniques, including both uninformed and probabilistic methods. Specifically, we use Breadth-First Search (BFS) and Depth-First Search (DFS) as examples of uninformed search strategies. Monte Carlo Tree Search (MCTS) is incorporated as a form of probabilistic search.

\item \textbf{Dynamic Programming (DP)}: Dynamic programming is applied to puzzles that exhibit overlapping subproblems and optimal substructure.

\item \textbf{Greedy Algorithm}: Greedy algorithms are employed in puzzles where locally optimal choices are expected to lead to globally optimal solutions or the search space is too large for other techniques, often reflecting strategies used in real-world scenarios.

\item \textbf{Other Methods}: These include other algorithms, such as backtracking and simulated annealing.
\end{itemize}

\begin{table*}[t]
\centering
\small
\addtolength{\tabcolsep}{-0.25em}
\resizebox{1.0\linewidth}{!}{
\begin{tabular}{lcccccccccccccccccc}
\toprule
\multirow{3}{*}{\textbf{Model}} &
\multicolumn{4}{c}{\textbf{Single-Player Det.}} & \multicolumn{4}{c}{\textbf{Single-Player Sto.}} & 
\multicolumn{4}{c}{\textbf{Two-Player Det.}} & \multicolumn{4}{c}{\textbf{Two-Player Sto.}} & 
\multicolumn{2}{c}{\textbf{Score}} \\
\cmidrule(lr){2-5}  \cmidrule(lr){6-9} \cmidrule(lr){10-13} \cmidrule(lr){14-17} \cmidrule(lr){18-19}
 & \multicolumn{2}{c}{\textbf{Easy}} & \multicolumn{2}{c}{\textbf{Normal}} & \multicolumn{2}{c}{\textbf{Easy}} & \multicolumn{2}{c}{\textbf{Normal}} & \multicolumn{2}{c}{\textbf{Easy}} & \multicolumn{2}{c}{\textbf{Normal}} & \multicolumn{2}{c}{\textbf{Easy}} & \multicolumn{2}{c}{\textbf{Normal}} & \multirow{2}{*}{\textbf{Avg.}} & \multirow{2}{*}{\textbf{Best}} \\
 \cmidrule(lr){2-3}  \cmidrule(lr){4-5}  \cmidrule(lr){6-7}  \cmidrule(lr){8-9}  \cmidrule(lr){10-11}  \cmidrule(lr){12-13}  \cmidrule(lr){14-15}  \cmidrule(lr){16-17} 
 & Avg. & Best & Avg. & Best & Avg. & Best & Avg. & Best & Avg. & Best & Avg. & Best & Avg. & Best & Avg. & Best \\
\midrule
Custom & 0.89 & -- & 0.83 & -- & 0.75 & -- & 0.80 & -- & 0.59 & -- & 0.75 & -- & 0.55 & -- & 0.72 & -- & 0.73 & -- \\
\midrule
Deepseek-R1 & \cellcolor{red!20}{0.33} & \cellcolor{red!20}{0.53} & \cellcolor{red!5}{0.25} & \cellcolor{red!5}{0.43} & \cellcolor{red!35}{0.54} & \cellcolor{red!35}{0.89} & \cellcolor{red!35}{0.51} & \cellcolor{red!35}{0.94} & \cellcolor{red!5}{0.65} & \cellcolor{red!5}{0.77} & 0.66 & \cellcolor{red!5}{0.82} & 0.46 & 0.52 & 0.53 & 0.61 & \cellcolor{red!20}{0.52} & {0.66} \\
o4-mini & \cellcolor{red!5}{0.30} & 0.42 & \cellcolor{red!35}{0.32} & \cellcolor{red!20}{0.52} & \cellcolor{red!5}{0.44} & \cellcolor{red!35}{0.89} & \cellcolor{red!20}{0.48} & \cellcolor{red!20}{0.93} & \cellcolor{red!35}{0.69} & \cellcolor{red!35}{0.85} & \cellcolor{red!20}{0.73} & \cellcolor{red!20}{0.85} & 0.49 & 0.52 & {0.57} & 0.60 & \cellcolor{red!35}{0.53} & \cellcolor{red!20}{0.73} \\
Gemini-2.5-pro & \cellcolor{red!35}{0.34} & \cellcolor{red!35}{0.68} & \cellcolor{red!20}{0.31} & \cellcolor{red!35}{0.65} & 0.36 & 0.69 & {0.36} & 0.76 & \cellcolor{red!20}{0.66} & \cellcolor{red!20}{0.82} & \cellcolor{red!35}{0.74} & \cellcolor{red!35}{0.94} & 0.47 & 0.52 & 0.47 & 0.54 & {0.50} & \cellcolor{red!35}{0.74} \\
QwQ-32B & 0.21 & 0.42 & 0.07 & 0.20 & \cellcolor{red!5}{0.48} & {0.85} & 0.31 & 0.81 & 0.59 & 0.68 & 0.34 & 0.58 & \cellcolor{red!5}{0.53} & 0.65 & 0.37 & 0.54 & 0.37 & 0.59 \\
grok-3-mini & 0.28 & \cellcolor{red!5}{0.46} & 0.20 & 0.38 & 0.22 & 0.64 & 0.12 & 0.65 & 0.59 & 0.75 &{0.68} & 0.78 & \cellcolor{red!20}{0.54} & \cellcolor{red!20}{0.71} & \cellcolor{red!20}{0.62} & \cellcolor{red!20}{0.76} & 0.43 & 0.65 \\
Deepseek-V3 & 0.22 & 0.40 & 0.16 & 0.34 & \cellcolor{red!35}{0.54} & \cellcolor{red!5}{0.86} & 0.35 & \cellcolor{red!20}{0.93} & 0.51 & 0.64 & 0.45 & 0.60 & 0.45 & 0.52 & 0.40 & 0.54 & 0.40 & 0.54 \\
GPT-4.1 & 0.24 & 0.43 & 0.28 & 0.48 & \cellcolor{red!5}{0.44} & \cellcolor{red!20}{0.88} & \cellcolor{red!5}{0.44} & \cellcolor{red!20}{ 0.93} & 0.63 & \cellcolor{red!5}{0.77} &\cellcolor{red!5}{ 0.72} & 0.81 & \cellcolor{red!35}{0.56} & \cellcolor{red!35}{0.73} & \cellcolor{red!35}{0.65} & \cellcolor{red!35}{0.81} & \cellcolor{red!5}{ 0.51} & \cellcolor{red!5}{0.72} \\
Qwen-2.5-VL-72B & 0.19 & 0.36 & 0.16 & 0.49 & 0.41 & 0.80 & 0.30 & \cellcolor{red!5}{0.86} & 0.52 & 0.69 & 0.55 & 0.68 & 0.45 & \cellcolor{red!5}{0.70} & 0.46 & \cellcolor{red!5}{0.73} & 0.40 & 0.55 \\
Llama-3.3-70B & 0.18 & 0.41 & 0.17 & 0.40 & 0.40 & 0.82 & 0.30 & 0.84 & {0.50} & 0.69 & 0.56 & 0.62 & {0.50} & {0.66} & \cellcolor{red!5}{0.60} & 0.64 & 0.41 & 0.60 \\
Gemma-3-27B & 0.22 & 0.43 & 0.22 & 0.41 & 0.35 & 0.79 & 0.32 & 0.82 & 0.45 & 0.59 & 0.51 & 0.61 & 0.46 & {0.66} & 0.47 & {0.65} & 0.38 & 0.51 \\
Phi-4-multimodal & 0.00 & 0.00 & 0.00 & 0.00 & 0.06 & 0.24 & 0.01 & 0.20 & 0.05 & 0.07 & 0.05 & 0.07 & 0.19 & 0.25 & 0.05 & 0.08 & 0.05 & 0.19 \\

\bottomrule
\end{tabular}
}
\caption{Code-based normalized scores.}
\label{tab:code_puzzle_scores}
\vspace{-1.5em}
\end{table*}

\subsection{Evaluation Protocols}
To gain a holistic view of a model’s problem-solving ability under distinct modes of interaction, we design the following two evaluation protocols.
\paragraph{Instruction-based Evaluation.} 
Single-player deterministic puzzles are evaluated using 10 randomly generated instances with fixed seeds from 1 to 10 to ensure reproducibility. For two-player deterministic puzzles, each model pair competes on 5 instances (seeds 1–5), with each match repeated twice while alternating the first player to account for first-mover advantage. All evaluations are conducted at two difficulty levels: \emph{easy} and \emph{normal}. Stochastic puzzles are excluded from this setting due to their inherent variance and the high cost of running enough instances to achieve statistically robust conclusions.

\paragraph{Code-based Evaluation.} 
Each foundation model is sampled 32 times per puzzle to generate code, following the prompt templates described in \S\ref{code_template}. The resulting programs are then executed to play the games. For deterministic puzzles, we follow the same evaluation protocol as in the instruction-based setting. For single-player stochastic puzzles, each generated program is evaluated over 100 runs (seeds 1 to 100) across both difficulty levels. For two-player stochastic puzzles, each program competes in 50 runs (seeds 1 to 50), alternating player roles in each match. 

\subsection{Evaluation Metrics}
\label{evaluation}

We employ two primary metrics to evaluate model performance: \textbf{Normalized Score} and \textbf{Elo Score}, both derived from raw scores.

\paragraph{Raw Score.} In single-player games, raw scores are either binary or continuous. Binary puzzles assign a score of 1 for success and 0 for failure. Continuous-score puzzles assign values based on criteria such as move count, constraints met, or objectives achieved, and scores may fall outside the [0, 1] range. In two-player games, outcomes are categorized as win, loss, or tie, corresponding to scores of 1, 0, and 0.5, respectively.

\paragraph{Normalized Score.} For two-player games, raw scores already lie in [0, 1] and do not require normalization. For single-player games, normalization ensures comparability by rescaling scores to the [0, 1] interval. This involves determining the best and worst achievable scores under identical initialization conditions. If higher scores are better, the top-performing model is assigned 1 and others receive $score/max$; if lower is better, normalization uses $min/score$.

\paragraph{Elo Score.} To enable unified comparison across both single-player and two-player settings, we apply the Elo rating system, a widely-used model comparison metric~\cite{boubdir-etal-2023-elo}. For single-player games, we create pairwise matchups between models based on their normalized scores—the model with the higher normalized score is considered the winner in each pairwise comparison. The implementation details are described in \S~\ref{imple_elo}.

\section{Experiments} \label{sec:exp}
\begin{table*}[ht]
\centering
\small
\addtolength{\tabcolsep}{-0.3em}
\resizebox{1.0\linewidth}{!}{%
\begin{tabular}{lcccccccccccc}
\toprule
\multirow{3}{*}{\textbf{Model}} &
\multicolumn{6}{c}{\textbf{SudoKillM}} &
\multicolumn{6}{c}{\textbf{SuperplyM}} \\
\cmidrule(lr){2-7} \cmidrule(lr){8-13}
& \multicolumn{3}{c}{\textbf{Easy}} & \multicolumn{3}{c}{\textbf{Normal}} & \multicolumn{3}{c}{\textbf{Easy}} & \multicolumn{3}{c}{\textbf{Normal}} \\
\cmidrule(lr){2-4} \cmidrule(lr){5-7} \cmidrule(lr){8-10} \cmidrule(lr){11-13}
& Score & vs.~Text & vs.~Custom & Score & vs.~Text & vs.~Custom & Score & vs.~Text & vs.~Custom & Score & vs.~Text & vs.~Custom \\
\midrule
Gemini-2.5-pro & 0.48 & \cellcolor{red!28}{0.21} & \cellcolor{red!2}{0.46} & 0.60 & \cellcolor{red!4}{0.56} & \cellcolor{green!34}{1.00} & 0.60 & \cellcolor{red!5}{0.54} & \cellcolor{green!34}{0.20} & 0.56 & \cellcolor{green!1}{0.57} & \cellcolor{red!22}{0.32} \\
Gemma-3-27B & 0.25 & \cellcolor{green!24}{0.37} & \cellcolor{red!50}{0.00} & 0.25 & \cellcolor{red!2}{0.24} & \cellcolor{red!50}{0.00} & 0.25 & \cellcolor{red!4}{0.23} & \cellcolor{red!40}{0.05} & 0.25 & \cellcolor{red!16}{0.17} & 0.25 \\
GPT-4.1 & 0.48 & \cellcolor{red!3}{0.45} & \cellcolor{red!19}{0.30} & 0.36 & \cellcolor{green!6}{0.40} & \cellcolor{red!36}{0.10} & 0.58 & \cellcolor{red!5}{0.53} & \cellcolor{red!25}{0.30} & 0.58 & \cellcolor{red!2}{0.56} & \cellcolor{red!33}{0.20} \\
o4-mini & 0.72 & \cellcolor{red!7}{0.62} & \cellcolor{red!36}{0.20} & 0.54 & \cellcolor{green!5}{0.59} & \cellcolor{red!9}{0.40} & 0.94 & \cellcolor{red!6}{0.83} & \cellcolor{green!2}{0.98} & 0.94 & \cellcolor{red!6}{0.83} & \cellcolor{red!2}{0.90} \\
Phi-4-multimodal & 0.20 & \cellcolor{red!25}{0.10} & \cellcolor{red!50}{0.00} & 0.35 & \cellcolor{red!30}{0.14} & \cellcolor{red!50}{0.00} & 0.25 & \cellcolor{red!32}{0.09} & \cellcolor{green!100}{1.00} & 0.27 & \cellcolor{red!6}{0.24} & \cellcolor{green!6}{0.30} \\
Qwen-2.5-VL-72B & 0.30 & \cellcolor{red!20}{0.18} & \cellcolor{red!34}{0.10} & 0.24 & \cellcolor{red!19}{0.15} & \cellcolor{red!50}{0.00} & 0.24 & \cellcolor{red!25}{0.12} & \cellcolor{green!13}{0.30} & 0.32 & \cellcolor{red!25}{0.16} & \cellcolor{red!3}{0.30} \\
\bottomrule
\end{tabular}%
}
\caption{Normalized score on \textsc{SudoKillM} and \textsc{SuperplyM} puzzles. Colors indicate the percentage change of the text-only or custom-strategy baseline versus the multimodal model score.}
\label{tab:multimodal}
\end{table*}

\subsection{Experimental Setup}

The foundation models we evaluate include GPT-4.1$^*$~\cite{gpt41}, o4-mini$^*$~\cite{o4}, Gemini-2.5-pro$^*$~\cite{gemini}, grok-3-mini$^*$~\cite{grok}, DeepSeek-V3~\cite{deepseekai2025deepseekv3technicalreport}, DeepSeek-R1~\cite{deepseekai2025deepseekr1incentivizingreasoningcapability}, QwQ-32B~\cite{qwq}, Qwen-2.5-VL-72B~\cite{bai2025qwen25vltechnicalreport}, Gemma-3-27B~\cite{gemmateam2025gemma3technicalreport}, Llama-3.3-70B~\cite{grattafiori2024llama3herdmodels}, and Phi-4-multimodal~\cite{microsoft2025phi4minitechnicalreportcompact}.%
\footnote{Models marked with an asterisk (*) are proprietary.} Models grok-3-mini, DeepSeek-V3, DeepSeek-R1, and QwQ-32B do not support image modalities and are therefore excluded from evaluation on text-image puzzles in the instruction-based setting. We use the chat or instruct versions of each model, as solving most puzzles involves multi-turn interactions.



\subsection{Main Results}

\autoref{tab:det_puzzle_scores} presents the normalized scores of all models under the instruction-based setting, while Elo scores are reported in \autoref{tab:det_puzzle_scores_updated} in Appendix. A breakdown of scores for each puzzle is provided in \S~\ref{more_results}. For the code-based setting, results are shown in \autoref{tab:code_puzzle_scores}.

\paragraph{Reasoning models outperform non-reasoning models in the instruction-based setting.}
From \autoref{tab:det_puzzle_scores}, we observe that reasoning models consistently outperform non-reasoning ones, with all top-5 models employing reasoning strategies. This demonstrates the effectiveness of test-time scaling using extended CoT, where deeper deliberation translates to better performance in gameplay. Notably, the relatively small QwQ-32B model surpasses larger non-reasoning models such as GPT-4.1 and DeepSeek-V3. Furthermore, open-source models are highly competitive with proprietary systems: for instance, DeepSeek-R1 achieves the highest normalized score of 0.62, outperforming Gemini-2.5-pro, the best-performing proprietary model, which scores 0.58. These findings indicate that open-source models are closing the performance gap. Although foundation models still lag behind our custom strategy (which scores 0.70) on average, several leading models perform comparably—or even better—in two-player deterministic puzzles, highlighting the rapid progress of foundation models.

\paragraph{Code-based setting is more challenging and leads to a notable performance drop.}
As shown in \autoref{tab:code_puzzle_scores}, model performance declines significantly in the code-based setting, where models must generate executable code to play the games autonomously. Unlike the instruction-based setting—where models act as interactive agents with ongoing access to game states and can adjust actions dynamically—the code-based setting demands strong program synthesis capabilities. This shift reduces the advantage of reasoning models: for example, GPT-4.1, a non-reasoning model, ranks among the top-3 performers in the code-based setting, whereas no non-reasoning model appears in the top-5 for the instruction-based setting.

The performance drop is especially evident in single-player deterministic puzzles. DeepSeek-R1, for instance, sees its score decline from 0.64 to 0.33 in the easy level, and from 0.48 to 0.25 in the normal level. \autoref{tab:code_instruction_comparison} in Appendix further reveals a significant reduction in win rates against the customized strategy across all models in the code-based setting in two-player deterministic games. Although the code-based setting underperforms compared to the instruction-based setting, its lower computational cost makes it a promising direction.

These results underscore the greater difficulty of the code-based setting, which not only tests reasoning but also code generation and execution accuracy. However, one advantage of this setting is efficiency: code is generated once per puzzle and can be reused. In our experiments, each model generates 32 samples per puzzle. As shown in \autoref{tab:code_puzzle_scores}, the best scores from code-based runs can approach or even match the performance of the customized strategy.

\begin{figure*}[!t]
    \centering
    \begin{subfigure}[b]{0.45\textwidth}
        \includegraphics[width=\linewidth]{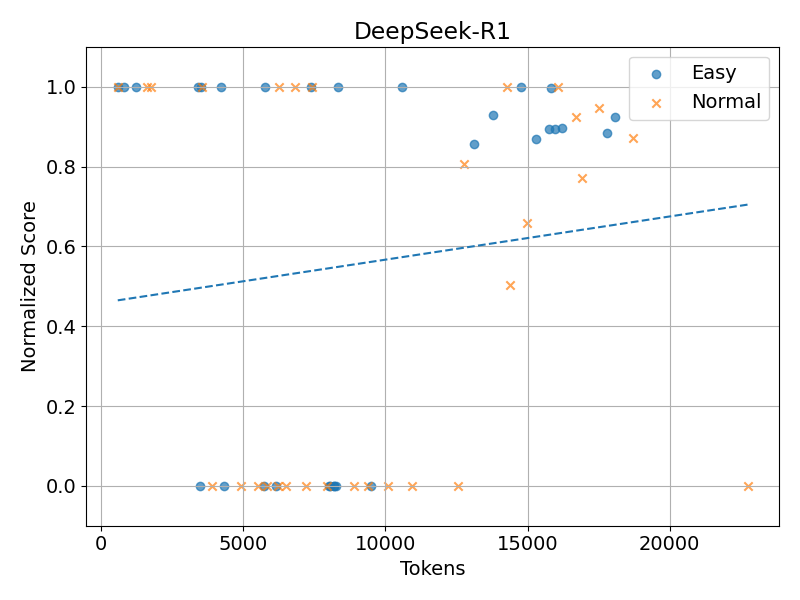}
        \label{fig:line1}
    \end{subfigure}
    \hfill
    \begin{subfigure}[b]{0.45\textwidth}
        \includegraphics[width=\linewidth]{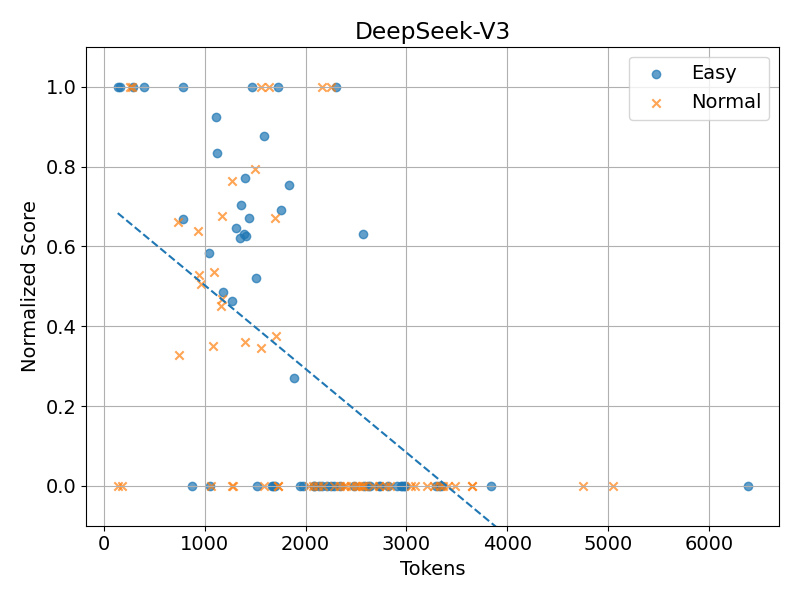}
        \label{fig:line2}
    \end{subfigure}
    \caption{Comparison between the reasoning model Deepseek-R1 and the non-reasoning model Deepseek-V3 in terms of generated token counts versus normalized scores on single-player deterministic puzzles.}
    \label{fig:two_graphs}
    \vspace{-1.5em}
\end{figure*}

\begin{table*}[ht]
\centering
\tiny
\addtolength{\tabcolsep}{-0.3em}
\resizebox{\linewidth}{!}{
\begin{tabular}{lccccccr}
\toprule
\multirow{2}{*}{\textbf{Model}} &
\multicolumn{6}{c}{\textbf{Status Type}} &
\multirow{2}{*}{\textbf{\#Token}} \\
\cmidrule(lr){2-7}
& Legal & Not Follow Instr. & Timeout & Rule Violation & Runtime Err. & Syntax Err. & \\
\midrule
\multicolumn{8}{c}{\emph{\textbf{Instruction-based}}} 
\\\noalign{\vskip 0.5ex}
Deepseek-R1 & 0.79 & 0.01 & -- & 0.20 & -- & -- & 9420.78 \tiny{$\pm$710.39} \\
o4-mini & 0.79 & 0.02 & -- & 0.19 & -- & -- & 4508.83 \tiny{$\pm$557.87} \\
Gemini-2.5-pro & 0.72 & 0.17 & -- & 0.12 & -- & -- & 12124.58 \tiny{$\pm$1016.97} \\
QwQ-32B & 0.78 & 0.01 & -- & 0.21 & -- & -- & 11840.75 \tiny{$\pm$835.78} \\
grok-3-mini & 0.82 & 0.04 & -- & 0.15 & -- & -- & 12479.65 \tiny{$\pm$976.22} \\
Deepseek-V3 & 0.72 & 0.03 & -- & 0.25 & -- & -- & 2013.47 \tiny{$\pm$95.00} \\
GPT-4.1 & 0.67 & 0.01 & -- & 0.32 & -- & -- & 1587.50 \tiny{$\pm$137.30} \\
Qwen-2.5-VL-72B & 0.57 & 0.05 & -- & 0.38 & -- & -- & 654.02 \tiny{$\pm$50.49} \\
Llama-3.3-7B & 0.61 & 0.00 & -- & 0.38 & -- & -- & 949.28 \tiny{$\pm$74.05} \\
Gemma-3-27B & 0.37 & 0.44 & -- & 0.19 & -- & -- & 1236.40 \tiny{$\pm$79.88} \\
Phi-4-multimodal & 0.42 & 0.17 & -- & 0.41 & -- & -- & 765.48 \tiny{$\pm$87.38} \\
\midrule
\multicolumn{8}{c}{\emph{\textbf{Code-based}}} 
\\\noalign{\vskip 0.5ex}
Deepseek-R1 & 0.54 & 0.18 & 0.01 & 0.00 & 0.11 & 0.16 & 11977.00 \tiny{$\pm$7694.16} \\
o4-mini & 0.61 & 0.26 & 0.03 & 0.03 & 0.05 & 0.03 & 1870.17 \tiny{$\pm$1379.59} \\
Gemini-2.5-pro & 0.58 & 0.13 & 0.02 & 0.14 & 0.13 & 0.00 & 14821.33 \tiny{$\pm$10064.59} \\
QwQ-32B & 0.19 & 0.07 & 0.00 & 0.01 & 0.19 & 0.54 & 10742.93 \tiny{$\pm$7226.60} \\
grok-3-mini & 0.56 & 0.19 & 0.03 & 0.05 & 0.13 & 0.04 & 14708.74 \tiny{$\pm$12639.34} \\
Deepseek-V3 & 0.26 & 0.61 & 0.00 & 0.03 & 0.07 & 0.03 & 1133.52 \tiny{$\pm$1235.67} \\
GPT-4.1 & 0.57 & 0.18 & 0.02 & 0.03 & 0.21 & 0.01 & 1287.01 \tiny{$\pm$1022.89} \\
Qwen-2.5-VL-72B & 0.46 & 0.13 & -- & 0.09 & 0.04 & 0.27 & 607.89 \tiny{$\pm$286.81} \\
Llama-3.3-7B & 0.46 & 0.20 & 0.02 & 0.11 & 0.15 & 0.06 & 741.86 \tiny{$\pm$347.07} \\
Gemma-3-27B & 0.43 & 0.35 & -- & 0.16 & 0.06 & 0.01 & 918.74 \tiny{$\pm$545.65} \\
Phi-4-multimodal & 0.00 & 0.50 & -- & 0.00 & 0.03 & 0.48 & 587.53 \tiny{$\pm$656.44} \\
\bottomrule
\end{tabular}
}
\caption{Distribution of status types and average tokens used per model in instruction-based and code-based settings.}
\label{tab:error_type}
\vspace{-1.5em}
\end{table*}

\subsection{More Instruction-based Analysis}
\paragraph{Mixed Effectiveness of Advanced Prompting Strategies.}

\autoref{tab:ablation} in Appendix reports the performance of GPT-4.1 and o4-mini on two puzzles, \textsc{TidyTower} and \textsc{SudoKill}, under various prompting strategies. Overall, the effectiveness of advanced prompting techniques is mixed. For instance, 1-shot prompting yields negligible improvement in both puzzles. ToT prompting helps in \textsc{TidyTower} but shows minimal benefit in \textsc{SudoKill}. Interestingly, prompting without history—i.e., omitting previous model inputs and outputs from the current prompt—leads to a substantial performance boost in \textsc{TidyTower}, outperforming ToT, especially considering ToT's much higher computational cost. 

This suggests that current models still struggle with multi-hop reasoning and reflection. Including past reasoning steps may inadvertently mislead the model rather than help it retrospect effectively. 

We also evaluate a legality-aware prompting strategy, where the model is explicitly provided with a list of legal candidate moves. This is motivated by the observation that many model failures stem from selecting illegal actions, which lead to immediate losses or premature termination. \autoref{tab:ablation} in Appendix shows that providing legal candidates consistently improves performance. Notably, the reasoning-focused model o4-mini benefits more from these prompting strategies than GPT-4.1.

\paragraph{Evaluating Multimodal Integration in Strategic Reasoning.}
\autoref{tab:multimodal} shows that most models benefit from incorporating visual inputs, confirming the value of image-based state representations in puzzle-solving tasks. High-capacity models like {o4-mini} and {GPT-4.1} achieve notable gains, with {GPT-4.1} improving its win rate by +0.38 on \textsc{SuperplyM} (Normal). However, weaker models such as {Phi-4-multimodal} struggle to utilize visual information effectively, sometimes exhibiting performance drops (e.g., -0.75 on \textsc{SuperplyM} Easy).
These results suggest that while visual information aids intuitive understanding, effective multimodal reasoning requires advanced fusion capabilities. The benefits are more pronounced in simpler tasks, whereas complex scenarios demand stronger cross-modal reasoning, which current models often lack.

\subsection{Scaling Analysis}

\paragraph{Reasoning models demonstrate better scaling between token count and performance.} 
\autoref{fig:two_graphs} compares Deepseek-R1 (reasoning) and Deepseek-V3 (non-reasoning) in terms of total generated tokens (reasoning + completion) versus normalized scores on single-player deterministic puzzles. These puzzles are all single-pass and do not involve multi-round interactions, making them suitable for such analysis. The results show that for Deepseek-R1, performance generally improves with increased token generation, suggesting effective test-time scaling. In contrast, Deepseek-V3 exhibits a flatter or even downward trend, indicating limited benefit from generating more tokens. Furthermore, Deepseek-R1 tends to allocate more tokens to normal-difficulty instances than to easy ones, aligning with task complexity, while Deepseek-V3 shows little variation across difficulty levels.

\paragraph{Reasoning models show improved performance in instruction-based settings but mixed results in code generation.}
\label{status}
For each run instance, we define several termination statuses. \textsc{Legal} means the game ends normally. \textsc{Rule violation} occurs when models make moves that violate the rules, causing game termination. \textsc{Not following instruction} indicates that foundation models fail to follow instructions properly; in instruction-based settings, this means the model generates data in a format that prevents the system from extracting moves; in code-based settings, it means the model fails to generate code meeting our requirements. \textsc{Timeout} is a status exclusive to code-based settings, indicating that the model-generated code exceeds our predetermined runtime limit, forcing the game to stop. \textsc{Syntax error}, also specific to code-based settings, occurs when the model generates code containing syntax errors. \textsc{Runtime error}, another code-based status, happens when code executes but fails during runtime due to errors such as index exceptions.

From Table \ref{tab:error_type}, which shows the distribution of status types and average token usage per model in two different settings, we observe that in instruction-based settings, most reasoning models consume significantly more tokens than non-reasoning models, with typical reasoning models using more than five times the tokens of their non-reasoning counterparts (though o4-mini is an exception with more modest token usage). In code-based settings, o4-mini's token usage remains similar to instruction-based settings, while other reasoning models consume substantially more tokens—approximately ten times that of non-reasoning models.

Regarding status types in instruction-based settings, reasoning models generally make fewer errors, suggesting that increased reasoning tokens at test-time correlate with error reduction. However, in code-based settings, the situation differs. While existing research demonstrates that large reasoning models excel in competitive programming \cite{openai2025competitiveprogramminglargereasoning}, our puzzle scenario yields different results. The table indicates that the best non-reasoning model, GPT-4.1, remains comparable to reasoning models, while one reasoning model, QwQ-32B, shows a notably low legal rate due to a high incidence of syntax errors in its code generation.



\section{Conclusion}

\ours is the first benchmark to compare reasoning techniques on puzzles that span text and vision modalities, deterministic and stochastic dynamics, and long-horizon interactions. It enables a systematic evaluation of models through both instruction-based and code-based settings.
We find that reasoning models perform best in instruction-based settings, benefiting from increased test-time computation. Open-source models such as DeepSeek-R1 match or surpass proprietary models, demonstrating rapid progress. In contrast, the code-based setting poses greater challenges due to the need for accurate program synthesis, though its lower computational cost and scalability make it a promising direction. Best-of-n sampling significantly improves performance in this setting. 
Multimodal inputs and legality-aware prompting offer further gains in specific scenarios. However, our analysis reveals that models often struggle with multi-hop reasoning—e.g., in \textsc{TidyTower}, removing prior reasoning history improves accuracy, suggesting that current models may be misled by irrelevant context. 
Overall, \ours offers a testbed for advancing reasoning and planning in foundation models, highlighting limitations of current systems and consequent opportunities for future research.

\section*{Limitations}
Although \ours spans 15 carefully curated puzzles, the overall number of puzzles remains modest, so results may be sensitive to the specific puzzle mix and random seeds. Moreover, due to rapid model evolution and budget constraints, our experiments may not include the latest model releases available after the experiment period. Finally, \ours does not yet assess whether fine-tuned models can outperform existing LLMs, which could provide additional insights.

\section*{Ethics Statement}
Our study uses only rule-based puzzles—no human subjects or personally identifiable information. All puzzles are original or permissively licensed. We execute model-generated code in a sandbox, follow provider terms/safety policies, and log only non-sensitive metadata.

\bibliography{references}

\clearpage
\appendix

\addtocontents{toc}{\protect\setcounter{tocdepth}{3}}

\hypersetup{linkcolor=black}

\onecolumn
\tableofcontents
\clearpage

\clearpage
\section{\ours}

\subsection{Dataset Overview}
\label{dataset}
\begin{table*}[!h]
\centering
\renewcommand{\arraystretch}{1.5} 
\setlength{\tabcolsep}{10pt} 
\rowcolors{2}{gray!15}{white} 

\resizebox{\linewidth}{!}{
\begin{tabular}{%
    >{\raggedright\arraybackslash}p{5cm} 
    >{\centering\arraybackslash}p{5cm} 
    >{\centering\arraybackslash}p{3cm} 
    >{\centering\arraybackslash}p{3cm} 
    >{\centering\arraybackslash}p{3cm}
}
\toprule
\textbf{Name} & \textbf{Scenario} & \textbf{Reward} & \textbf{Data} & \textbf{Main Reasoning}  \\
\midrule

SudoKill & Two-player & Deterministic & Text & Logical, Spatial  \\
TidyTower \cite{shasha2023tidy} & Single-player & Deterministic & Text & Spatial \\
CardNim \cite{shasha2022card} &  Two-player & Deterministic & Text & Numerical, Logical  \\
OptimalTouring & Single-player & Deterministic & Text & Numerical\\
CountMaximalCocktails \cite{shasha2022maximal} & Single-player & Deterministic & Text & Logical \\
MaxMaximalCocktails &  Two-player & Deterministic & Text & Logical \\
ExclusivityParticles \cite{shasha2022exclusivity} &  Two-player & Deterministic & Text & Numerical, Spatial  \\
ExclusivityProbes & Single-player & Stochastic & Text & Numerical, Spatial \\
RubyRisks \cite{shasha2017ruby} & Single-player & Stochastic & Text & Numerical, Logical\\
BeatOrBombSto. &  Two-player & Stochastic & Text & Logical, Numerical  \\
MaxTarget & Single-player & Stochastic & Text & Logical, Numerical \\
LargerTarget &  Two-player & Stochastic & Text & Logical, Numerical \\
Superply &  Two-player & Deterministic & Text & Numerical, Spatial \\
SudoKill M. &  Two-player & Deterministic & Text-Image & Visual, Logical  \\
Superply M. &  Two-player & Deterministic & Text-Image & Visual, Numerical \\
\bottomrule
\end{tabular}
}
\caption{Overview of Puzzle Games.}
\label{tab:puzzleplex_detailed_table}
\end{table*}

\clearpage
\subsection{Example of Simulator}
\label{simulator}

\begin{figure*}[ht]
    \centering
    \begin{minipage}{0.48\textwidth}
        \centering
        \includegraphics[width=\linewidth]{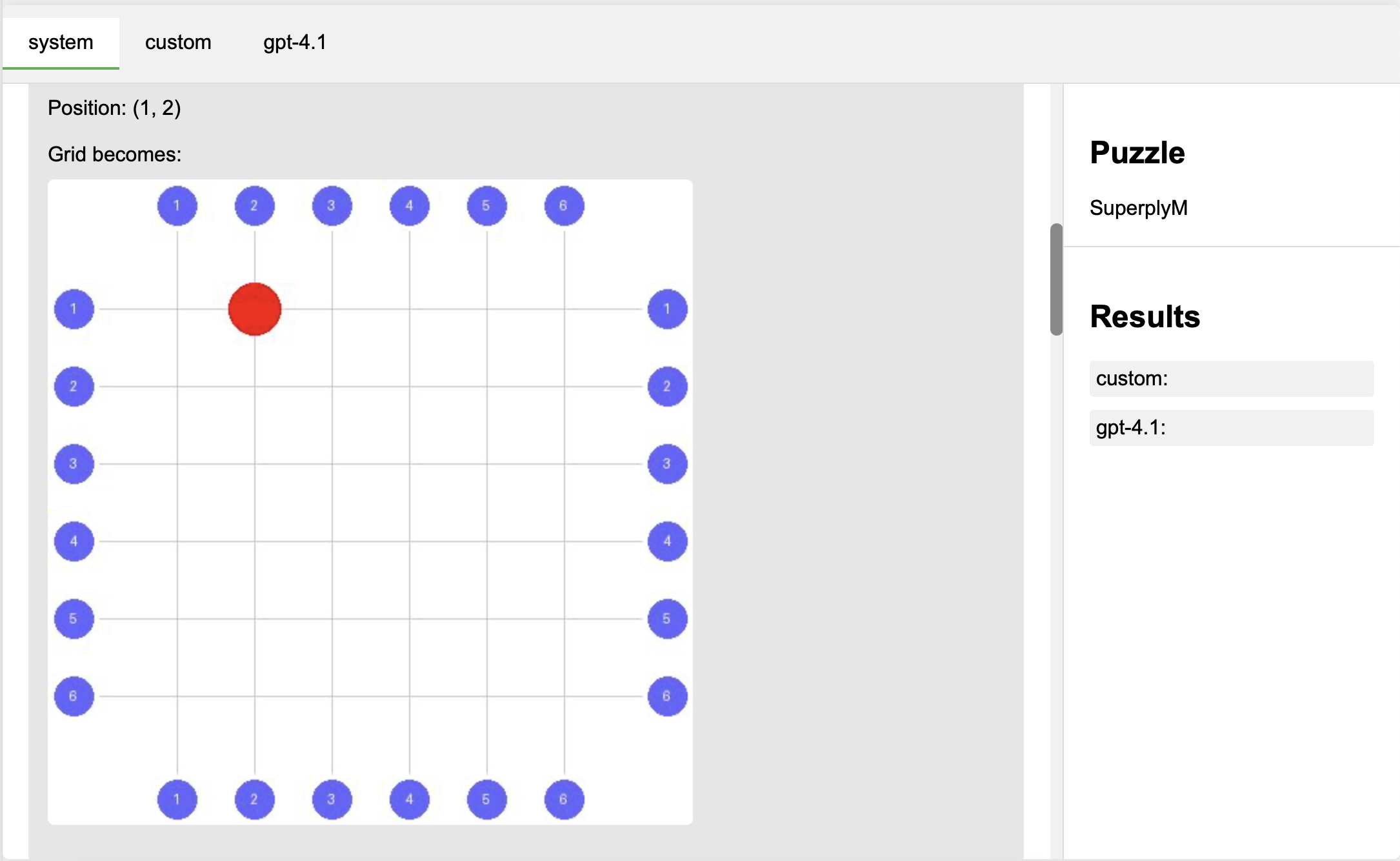}
        \caption*{(a)}
    \end{minipage}
    \hfill
    \begin{minipage}{0.48\textwidth}
        \centering
        \includegraphics[width=\linewidth]{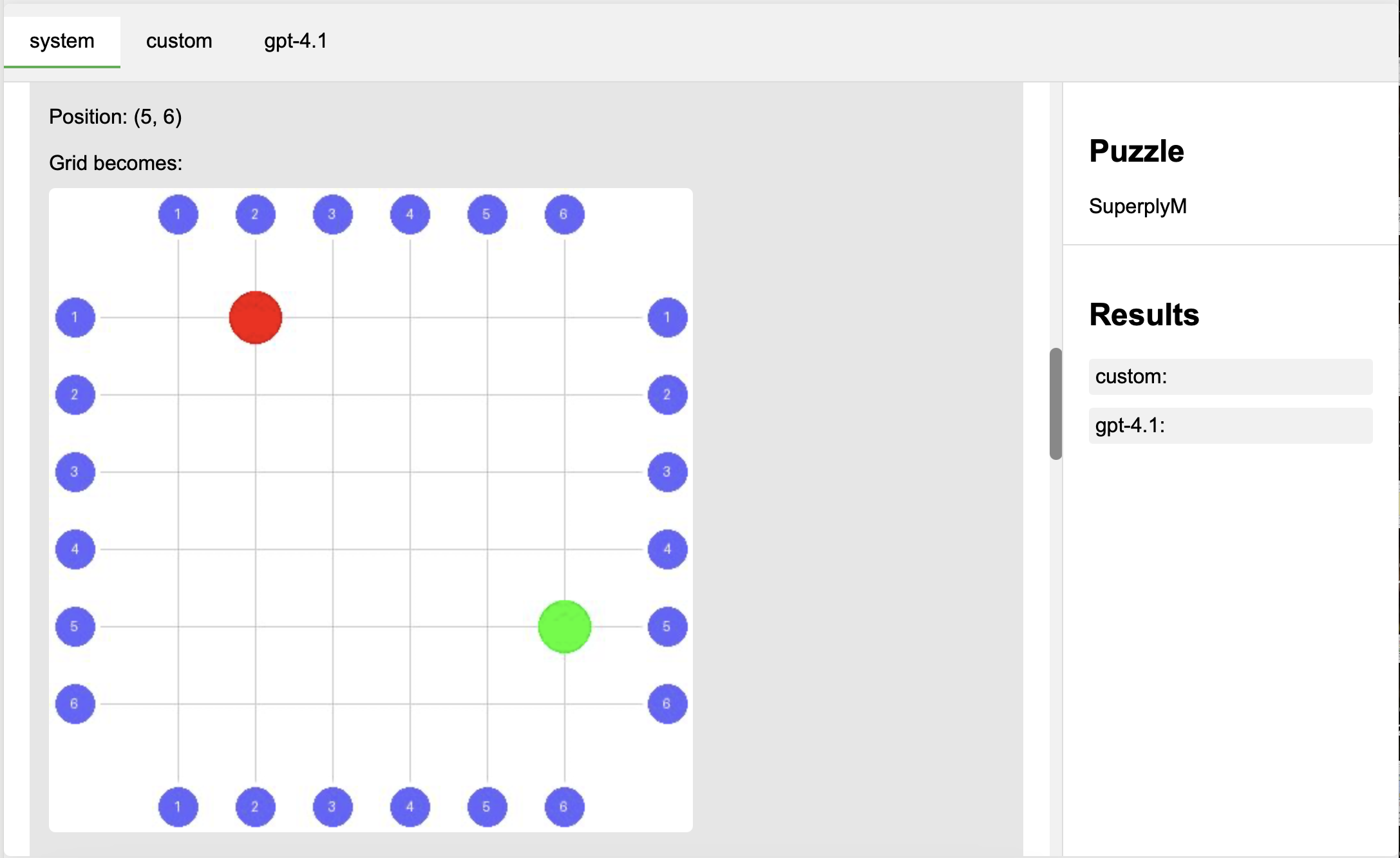}
        \caption*{(b)}
    \end{minipage}
    
    \vspace{0.5cm}
    
    \begin{minipage}{0.48\textwidth}
        \centering
        \includegraphics[width=\linewidth]{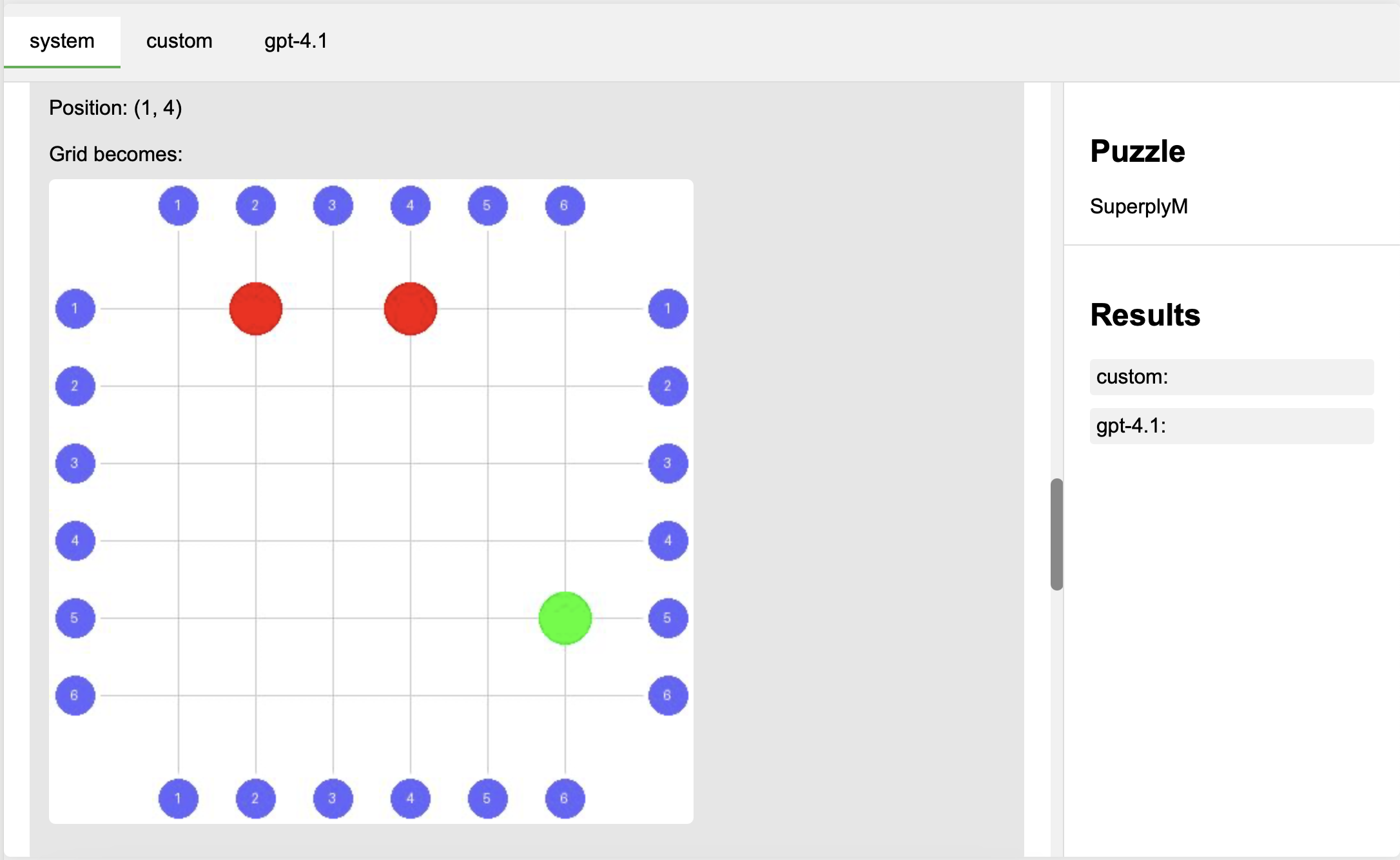}
        \caption*{(c)}
    \end{minipage}
    \hfill
    \begin{minipage}{0.48\textwidth}
        \centering
        \includegraphics[width=\linewidth]{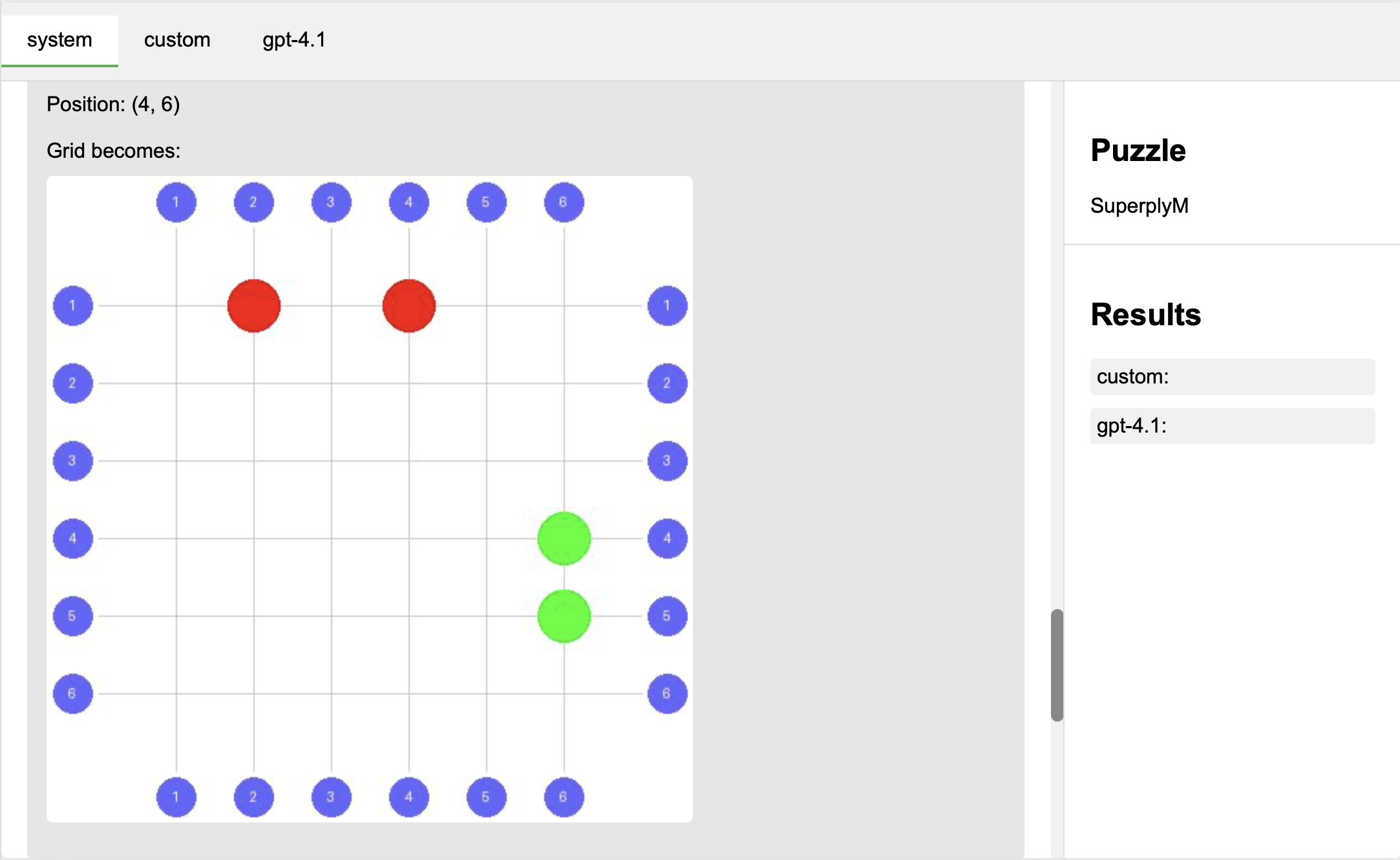}
        \caption*{(d)}
    \end{minipage}
    
    \caption{Overview of \textbf{Simulator}.
    The purpose of the Simulator is to present a history of the moves for a given puzzle for review by people. The representation of that history will differ for each kind of puzzle and the particular steps will depend on the methods used. SuperplyM is a two-player puzzle whose pedagogical goal for people is to teach arithmetic (e.g. multiplication). Play alternates between red and green players. When the red player responds correctly to a hint (panels a and c), the location chosen by that player turns red. When the green player responds (panels b and d), the location chosen by the green player turns green. In this image, we see a sequence of four moves, two by red and two by green, illustrating the history of moves taken by each method. }
    \label{fig:simulator}
    
\end{figure*}

\clearpage
\subsection{Breakdown Description of Puzzles}
\label{description}
\begin{figure*}[h!]
\begin{tcolorbox}[colback=white, colframe=blue!50!black, title=SudoKill, fontupper=\footnotesize, fonttitle=\footnotesize]
\textbf{Rule} \\
Sudokill is a competitive two-player variant of the classic Sudoku game. Like standard Sudoku, the game is played on a grid where the objective is to fill each row, column, and subgrid with the numbers from 1 to n, where n is the size of the row or column, without repeating any number in the same row, column, or subgrid.\\

In Sudokill, players take turns placing a number into an unoccupied cell. The first player can choose any empty cell to start the game. After that, each player must place their number in an unoccupied cell that lies in the \textbf{same row or column} as the last move made by their opponent. If there are no such cells available, the player may choose any unoccupied cell on the board. \\

A move is considered \textbf{invalid} if it violates standard Sudoku rules (i.e., placing a number that already appears in the same row, column, or subgrid), or if it is made in a cell not allowed by the row/column constraint described above. The first player to make an invalid move \textbf{loses} the game.\\

\textbf{Example} \\
If the current grid is 
\begin{verbatim}
[6, 8, 4, 5, 1, 3, 2, 7, 9],
[5, 9, 7, 6, 2, 0, 1, 8, 0],
[2, 3, 1, 4, 8, 7, 6, 5, 0],
[9, 1, 2, 7, 6, 4, 8, 0, 3],
[4, 6, 8, 3, 0, 1, 7, 2, 5],
[7, 5, 3, 2, 9, 8, 4, 1, 6],
[8, 4, 5, 1, 3, 2, 9, 6, 7],
[1, 0, 6, 9, 0, 5, 0, 3, 8],
[3, 2, 0, 0, 7, 0, 5, 4, 0]
\end{verbatim}

and now is your turn and the previous move by the opponent is to fill the cell at (0, 8) with the value 9. So now the cells you can place a number are [(1,8), (2,8), (8,8)] because you can only place a number in the same row or column as the last move. \\

If the current grid is
\begin{verbatim}
[6, 8, 4, 5, 1, 3, 2, 7, 9],
[5, 9, 7, 6, 2, 0, 1, 8, 0],
[2, 3, 1, 4, 8, 7, 6, 5, 0],
[9, 1, 2, 7, 6, 4, 8, 0, 3],
[4, 6, 8, 3, 0, 1, 7, 2, 5],
[7, 5, 3, 2, 9, 8, 4, 1, 6],
[8, 4, 5, 1, 3, 2, 9, 6, 7],
[1, 0, 6, 9, 0, 5, 0, 3, 8],
[3, 2, 0, 0, 7, 0, 5, 4, 1]
\end{verbatim}

and now is your turn and the previous move by the opponent is to fill the cell at (0, 8) with the value 9. Now you can fill the cell (1, 8) with the value 4 to win this game because after you fill the cell (1, 8) with the value 4, the opponent can only fill the cell (2, 8) and (1, 5), but no matter which value the opponent fills in these two cells will violate the rules. 
\end{tcolorbox}
\caption{Description of SudoKill.}
\end{figure*}
\begin{figure*}[h!]
\begin{tcolorbox}[colback=white, colframe=blue!50!black, title=TidyTower, fontupper=\footnotesize, fonttitle=\footnotesize]
\textbf{Rule} \\
Tidy Tower is a single-player puzzle involving a vertical stack of cubes, each with four colored sides arranged in a fixed clockwise order: \textbf{Red} (R), \textbf{Yellow} (Y), \textbf{Blue} (B), and \textbf{Green} (G). The player's objective is to transform the tower such that all cubes display the same color on their front face—this state is referred to as a tidy tower. \\

Two types of operations are allowed to manipulate the tower: \\

\textbf{Rotate}: When a player rotates a cube at a certain index, that cube and all cubes above it rotate together in a clockwise direction. A single rotation shifts the front-facing side of a cube to the next color in the clockwise sequence. For example, rotating once changes a cube with front face "R" to "Y", and so on. Rotating four times returns it to the original orientation. \\

\textbf{Rotate with Holding}: A player can also rotate a cube while holding a cube above it. This operation rotates only the selected cube and all cubes below it, while the held cube and any cubes above it remain in place. \\

\textbf{Example} \\
The initial setting is: RGBYRGBYBGBGBG. Can you make this tower tidy in eight moves or less?
Solution for eight moves: \\
RGBYRGBYBGBGBG → (rotate by one position at position 1 and not hold at position 2) \\
RRGBYRGBGRGRGR → (rotate by one position at position 2 and hold at position 3) \\
RRRBYRGBGRGRGR → (rotate by two positions at position 3 and hold at position 4) \\
RRRRYRGBGRGRGR → (rotate by one at position 4 and hold at position 5) \\
RRRRRRGBGRGRGR → (rotate by one at position 6 and hold at position 9) \\
RRRRRRRGRRGRGR → (rotate by one at position 7 and hold at position 8) \\
RRRRRRRRRRGRGR → (rotate by one at position 10 and hold at position 11) \\
RRRRRRRRRRRRGR → (rotate by one at position 12 and hold at position 13) \\
RRRRRRRRRRRRRR → Done

\end{tcolorbox}
\caption{Description of TidyTower.}
\end{figure*}
\begin{figure*}[h!]
\begin{tcolorbox}[colback=white, colframe=blue!50!black, title=CardNim, fontupper=\footnotesize, fonttitle=\footnotesize]
\textbf{Rule} \\
Card Nim is a two-player turn-based game played with a shared pile of stones and individual hands of number cards. At the start of the game, both players receive a set of cards, each card displaying a positive integer. A single pile of stones is placed at the center of the board. \\

On each turn, a player must play one of their cards to remove exactly that number of stones from the pile. A card can only be played if its value is \textbf{less than or equal} to the number of stones remaining. Once a card is used, it cannot be reused. The two players take turns alternately. \\

The objective is to be the player who removes the \textbf{last stone} from the pile. However, if a player is unable to play any card on their turn—because all of their remaining cards are greater than the number of remaining stones—they \textbf{lose the game} immediately. \\

\textbf{Example} \\
For example, suppose there are five stones left and each of the two players you and your opponent has three cards with 1, 2,and 3, respectively. You goes first. Who wins? \\Your opponent wins. If you removes 2 or 3, then opponent can win immediately with 3 or 2 respectively.\\
So, you removes 1. Now your opponent removes 3, leaving 1. Now you has only cards with numbers greater than 1 so you lose.
\end{tcolorbox}
\caption{Description of CardNim.}
\end{figure*}

\begin{figure*}[h!]
\begin{tcolorbox}[colback=white, colframe=blue!50!black, title=OptimalTouring, fontupper=\footnotesize, fonttitle=\footnotesize]
\textbf{Rule} \\
Optimal Touring is a route optimization puzzle in which a player must plan a one-day tour across a set of tourist sites. Each site is defined by the following attributes: \\
- A \textbf{location} represented by street and avenue coordinates. \\
- A \textbf{fixed visiting time} (in minutes) that must be spent at the site. \\
- A \textbf{value}, indicating the reward or importance of visiting the site. \\
- A \textbf{visiting window} specified by a start and end hour (in 24-hour format), representing when the site is accessible. \\

The player’s objective is to \textbf{maximize the total value} of visited sites while adhering to time and location constraints. \\

The total time spent on the tour includes: 

- The \textbf{visit time} required at each site. 

- The \textbf{travel time} between consecutive sites, computed using \textbf{Manhattan distance} (i.e., the sum of the absolute differences in street and avenue numbers). \\

The tour can start at \textbf{any site}, but each site must be visited \textbf{within its allowed time window}, and the cumulative time (including both visiting and travel) must respect this schedule. Once a site is visited, its value is counted toward the total. \\

\textbf{Example} \\ 
Here is an example of data: \\
\begin{tabular}{ccccccc}
\toprule
Site & Avenue & Street & Desired Time & Value & Begin Hour & End Hour \\
\midrule
1 & 50 & 96 & 114 & 3 & 6 & 12 \\
2 & 8 & 23 & 190 & 186 & 9 & 17 \\
3 & 88 & 69 & 218 & 3 & 9 & 12 \\
4 & 0 & 95 & 101 & 86 & 6 & 12 \\
5 & 1 & 48 & 192 & 199 & 5 & 12 \\
\bottomrule
\end{tabular} 

If you start visit cite 5 at 5:00, then go to site 2, then the hour is after 12:00, and the value you get is 199 + 186 = 385.
\end{tcolorbox}
\caption{Description of OptimalTouring.}
\end{figure*}

\begin{figure*}[h!]
\begin{tcolorbox}[colback=white, colframe=blue!50!black, title=CountMaximalCocktails, fontupper=\footnotesize, fonttitle=\footnotesize]
\textbf{Rule} \\
Count Maximal Cocktails is a combinatorial puzzle inspired by drug treatment for orphan diseases, where the objective is to discover safe and effective drug combinations. Each drug is represented as a \textbf{node} in an undirected graph. Pairs of drugs that should not be combined due to harmful interactions are represented as \textbf{edges} connecting the respective nodes. \\

A \textbf{cocktail} is a subset of drugs that can be administered together safely—meaning no two drugs in the subset have a harmful interaction. In graph theory, such a subset corresponds to an \textbf{independent set}, where no two nodes are directly connected by an edge. \\

The goal is to identify \textbf{all maximal independent sets} of the graph, referred to in this context as \textbf{maximal cocktails}. A set is \textbf{maximal} if it is an independent set and no additional drug can be added to it without introducing a harmful interaction (i.e., violating independence). Note that "maximal" does not mean "maximum in size"; rather, it means that the set cannot be extended further while maintaining its validity. \\

Two difficulty settings are defined:

In the \textbf{easy} level, only the number of maximal cocktails needs to be determined.

In the \textbf{normal} level, the task is to list all maximal cocktails explicitly. \\

The input includes a list of \textbf{drugs (nodes\_list)} and a \textbf{list of harmful interactions (edges\_list)}, where each edge is a pair of drug identifiers indicating a conflict. \\

\textbf{Example} \\
Suppose the drug list is [1, 2, 3, 4] and the bad interaction list is [(1, 2)]. \\

The maximal cocktails are [1, 3, 4] and [2, 3, 4], and the number of maximal cocktails is 2.
\end{tcolorbox}
\caption{Description of CountMaximalCocktails.}
\end{figure*}
\begin{figure*}[h!]
\begin{tcolorbox}[colback=white, colframe=blue!50!black,  title=MaxMaximalCocktails, fontupper=\footnotesize, fonttitle=\footnotesize]
\textbf{Rule} \\
Max Maximal Cocktails is a strategic two-player game played on a graph where nodes represent \textbf{drugs}, and edges represent \textbf{harmful interactions} between drug pairs. The game's core objective is to manipulate the structure of the graph by \textbf{adding edges}, while maintaining or increasing the number of valid drug combinations—called \textbf{maximal cocktails}. \\

A maximal cocktail is defined as a \textbf{maximal independent set} in the graph: a set of drugs in which no two drugs have a harmful interaction, and to which no more drugs can be added without creating a conflict. \\

At the beginning of the game, a list of nodes (drugs) is provided with \textbf{no edges}, meaning all combinations are potentially valid. Players take turns, and on each turn, a player \textbf{adds an edge} between two distinct nodes. The added edge represents the discovery or introduction of a harmful interaction between the two corresponding drugs. \\

The key rule is that \textbf{a move is only legal if it does not decrease} the current number of maximal cocktails. The first player who adds an edge that causes a decrease in the number of maximal cocktails loses the game. \\

\textbf{Example} \\
Suppose the node list is {[1, 2, 3]}, and you are the first player, you can add the edge {(1, 2)}, then the number of maximal cocktails is 2, which is larger than the number of maximal cocktails without the edge {(1, 2)}, which is 1. So this addition is legal. \\

But if your opponent adds the edge {(2, 3)} after you add the edge {(1, 2)}, then the number of maximal cocktails is 3, which is also legal. \\

After that, you will lose since you cannot add any edge to increase the number of maximal cocktails. 
\end{tcolorbox}
\caption{Description of MaxMaximalCocktails.}
\end{figure*}
\begin{figure*}[h!]
\begin{tcolorbox}[colback=white, colframe=blue!50!black,  title=ExclusivityParticles, fontupper=\footnotesize, fonttitle=\footnotesize]
\textbf{Rule} \\
Exclusivity Particles is a two-player combinatorial game played in a \textbf{d-dimensional binary space}, often conceptualized as the vertices of a \textbf{d-dimensional hypercube}. Each coordinate in this space is binary—either 0 or 1—representing discrete states along each dimension (e.g., spin up or down). \\

Players take turns placing \textbf{particles} at positions in this space. Each particle occupies a unique vertex of the hypercube. A strict exclusion principle governs the game: Any two particles must differ in at least \textbf{k dimensions}, meaning their \textbf{Hamming distance} must be \textbf{greater than or equal to k}. The Hamming distance is calculated as the number of differing coordinates between two binary vectors. \\
 
The game proceeds as follows:

- The \textbf{first player} places a particle at any position in the d-dimensional binary space.

- The \textbf{second player} then places another particle at a different position that satisfies the minimum distance condition with respect to all previously placed particles.

- Players alternate turns.

- A player \textbf{loses} if they cannot place a new particle that maintains the required distance \textbf{from all previously placed particles}.\\

\textbf{Example} \\
If the dimension is 3 and the required distance is 2, and you are the first player, you could place the first particle at [0, 0, 0]. \\
 
The second player could then place the second particle at [0, 1, 1]. If you place the third particle at [1, 0, 1], the second player cannot place a fourth particle that satisfies the condition and would lose.

\end{tcolorbox}
\caption{Description of ExclusivityParticles.}
\end{figure*}
\begin{figure*}[h!]
\begin{tcolorbox}[colback=white, colframe=blue!50!black,   title=ExclusivityProbes, fontupper=\footnotesize, fonttitle=\footnotesize]
\textbf{Rule} \\
Exclusivity Probes is a deductive search game played in a \textbf{d-dimensional binary space}, conceptually represented as a \textbf{d-dimensional hypercube}, where each position is a binary vector of length d (each dimension having value 0 or 1). \\

There are exactly num\_particles particles hidden in this space, and they obey a strict \textbf{exclusion principle}: any two particles must differ in at least k dimensions, meaning their \textbf{Hamming distance} must be greater than or equal to k. The Hamming distance between two positions is the number of coordinates in which they differ.\\

The player interacts with the environment by making probes. A probe is a query at a specific position in the hypercube. The response will be:

- "yes" if there is a particle at that exact position,

- "no" otherwise. \\

The objective is to identify the exact locations of all particles using \textbf{as few probes as possible}. \\

\textbf{Example} \\
If the dimension is 2, the number of particles is 2, and the distance is 1. \\

You can probe the position [0, 0], and if the response is 'yes', we only need one more probe to find the other particle because the particles can be either at locations [0, 0] and [1, 1] or at [0, 1] and [1, 0]. \\

If the response is 'no', we may need 3 more probes to find all the particles.
\end{tcolorbox}
\caption{Description of ExclusivityProbes.}
\end{figure*}
\begin{figure*}[h!]
\begin{tcolorbox}[colback=white, colframe=blue!50!black, title=RubyRisks, fontupper=\footnotesize, fonttitle=\footnotesize]
\textbf{Rule} \\
Ruby Risks is a sequential deduction game involving a set of num\_boxes hidden containers, each holding an \textbf{unknown number of identical rubies}. The total number of rubies across all boxes is known in advance and given as total\_rubies. \\

Each turn, the player submits a \textbf{single request}: a number of rubies they wish to take from \textbf{the next unopened box}. Boxes are opened in fixed left-to-right order, one per turn. \\

The outcome of a request depends on the number of rubies hidden in the box:

- If the request is \textbf{less than or equal} to the number of rubies in the box, the player \textbf{successfully collects} that amount.

- If the request is \textbf{greater than} the number of rubies in the box, the request fails, and the player receives \textbf{nothing} from that box. \\

The game proceeds turn by turn, with each turn corresponding to a new box. The goal is to \textbf{maximize the total number of rubies collected} across all turns. \\

\textbf{Example} \\
Suppose there are 3 boxes, and the hidden rubies in each box are: [11, 9, 10]. Total rubies = 30. \\

Turn 1: You request 10 rubies.

Feedback: 10 (successfully take 10 rubies). \\

Turn 2: You request 8 rubies.

Feedback: 8 (successfully take 8 rubies). \\

Turn 3: You request 12 rubies.

Feedback: 0 (because 12 > 10, so you get nothing from that box). \\

Total rubies collected so far: 18.
\end{tcolorbox}
\caption{Description of RubyRisks.}
\end{figure*}
\begin{figure*}[h!]
\begin{tcolorbox}[colback=white, colframe=blue!50!black, title=BeatOrBombSto, fontupper=\footnotesize, fonttitle=\footnotesize]
\textbf{Rule} \\
Beat Or Bomb Sto. is a two-player card game where players tactically choose when to \textbf{compete} or \textbf{give up} with the cards in their hands to maximize their total score across several rounds. \\

At the beginning, each player is given a set of num\_cards, which may differ in composition but are balanced so that the \textbf{total value} of each set is the same. Card values are assigned as follows:

- Numeric cards (2–10) are worth their face value.

- Jacks, Queens, Kings, and Aces have values of 11, 12, 13, and 1, respectively. \\

Each round proceeds as follows:

- Both players simultaneously select one card from their remaining set and decide whether to compete with it or give it up.

- This decision is \textbf{private}—neither player knows the other’s card or choice until both have confirmed.

- Regardless of the choice, the selected card is \textbf{removed} from the player’s hand. \\

Scoring rules:

- If both players compete, the player with the \textbf{higher} card earns points equal to \textbf{both card values combined}. The other player earns nothing.

- If both players give up, \textbf{no points are awarded}.

- If one player competes while the other gives up, the competing player earns points \textbf{equal to their card’s value}. The giving-up player earns nothing. \\

The game continues until all cards are used. The player with the \textbf{most points} at the end wins. \\

\textbf{Example} \\
In one of the rounds, if you choose to play the card '5' and compete, your opponent plays the card 'K' and give up, you will get 5 points and your opponent gets nothing.

\end{tcolorbox}
\caption{Description of BeatOrBombSto.}
\end{figure*}

\begin{figure*}[h!]
\begin{tcolorbox}[colback=white, colframe=blue!50!black,  title=MaxTarget, fontupper=\footnotesize, fonttitle=\footnotesize]
\textbf{Rule} \\
Max Targetis a probabilistic decision-making game where the player must choose bags of coins over a fixed number of turns to maximize the \textbf{total value of collected coins}. The game consists of bag\_count bags, each containing coins with specific known values (e.g., [1, 2], [3, 4]), but \textbf{the order of the bags is randomized} before gameplay begins. \\

At the start, the player is informed of:

- The list of coin values contained in each bag.

- The \textbf{total number of picks} allowed during the game (max\_guess). \\

Each turn proceeds as follows:

- The player chooses a \textbf{bag index}.

- One \textbf{random coin} is drawn from the chosen bag and added to the player's total score.

- The drawn coin is then \textbf{removed} from that bag.

- Over time, based on the drawn coins, the player can infer which observed bag maps to which known configuration. \\

\textbf{Example} \\
If you're told the bags contain [1, 2] and [3, 4], and the total number of picks is 2. \\

If you pick bag 0 and get a coin value of 4, then in the next turn, you will know that bag 0 contains [3, 4] and bag 1 contains [1, 2], and value 4 in bag 0 is removed and remaining values are [3]. \\

So, if you pick bag 0 again, you will get a coin value of 3, which is bigger than the coin value of bag 1. So, you should pick bag 0 again to maximize your score.
\end{tcolorbox}
\caption{Description of MaxTarget.}
\end{figure*}

\begin{figure*}[h!]
\begin{tcolorbox}[colback=white, colframe=blue!50!black,  title=LargerTarget, fontupper=\footnotesize, fonttitle=\footnotesize]
\textbf{Rule} \\
Larger Target is a two-player competitive coin-picking game. There are bag\_count bags, each containing a known list of coin values, but the order of the bags is \textbf{randomized} at the start of the game. Players alternate turns, each making a total of max\_guess picks across the game. \\

Each turn proceeds as follows:

- The player selects a \textbf{bag index}.

- A \textbf{random coin} is drawn from the chosen bag and removed from it.

- The value of the drawn coin is added to the player’s total score. \\

The objective is to \textbf{accumulate a higher total coin value} than the opponent by making informed decisions about which bag to choose. Since the order of bags is shuffled, players must deduce which real bag corresponds to each index by observing the coins drawn—both by themselves and their opponent. \\

\textbf{Example} \\
If you're told the bags contain [1, 2] and [3, 4], and the total number of picks is 2. \\

If your opponent pick bag 0 and get a coin value of 3, then in your turn, you will know that bag 0 contains [3, 4] and bag 1 contains [1, 2], and value 3 in bag 0 is removed and remaining values are [4]. So, if you pick bag 0 again, you will get a coin value of 4, which is bigger than the coin value of bag 1. So, you should pick bag 0 to make your score higher than your opponent.

\end{tcolorbox}
\caption{Description of LargerTarget.}
\end{figure*}
\begin{figure*}[h!]
\begin{tcolorbox}[colback=white, colframe=blue!50!black,  title=Superply, fontupper=\footnotesize, fonttitle=\footnotesize]
\textbf{Rule} \\
Superply is a two-player competitive path-building game played on a 1-indexed grid-based board. The grid is initially filled with zeros and players take turns selecting valid positions based on system-provided \textbf{mathematical hints}. \\

Each player has a unique path-building objective:

- Player 1 aims to build a continuous path of their claimed squares (marked with value 1) from the \textbf{left} edge of the grid to the \textbf{right}.

- Player 2 aims to build a path (marked with value 2) from the \textbf{top} edge to the \textbf{bottom}. \\

A valid path is a sequence of adjacent same-value cells, where adjacency includes both \textbf{sidewise and diagonal} (corner) neighbors. \\

Each turn proceeds as follows:

- The system provides a \textbf{hint}, such as a condition on the sum or product of the row and column indices (e.g., "sum < 10", "product contains digit 6").

- The player selects a grid cell (row, column) that: 1) Is currently \textbf{unoccupied} (i.e., value is 0); 2) \textbf{Satisfies the hint condition}.

- If the selection is \textbf{valid}, the cell is \textbf{updated} to reflect the player’s value (1 or 2). Otherwise, the move is \textbf{skipped and the turn passes} to the opponent. \\

The game ends when a player successfully builds a full path satisfying their objective. The first to do so wins. \\

\textbf{Example} \\
If the hint is "product contains digit 6," and the grid is as follows: 
\begin{verbatim}
[0, 0, 0, 0, 0, 0],
[0, 0, 0, 0, 0, 0],
[0, 0, 0, 0, 0, 0],
[0, 0, 0, 0, 0, 0],
[0, 0, 0, 0, 0, 0],
[0, 0, 0, 0, 0, 0]
\end{verbatim}
\vspace{1em}
If you are Player 1, you can select the position (1, 6), (6, 1), (2, 3), (3, 2) or (6, 6) because the product of the row and column indices is 6, 6, 6, 6 and 36, respectively, and they all contain the digit 6. \\

If you choose the position (6, 6), the grid becomes:
\begin{verbatim}
[0, 0, 0, 0, 0, 0],
[0, 0, 0, 0, 0, 0],
[0, 0, 0, 0, 0, 0],
[0, 0, 0, 0, 0, 0],
[0, 0, 0, 0, 0, 0],
[0, 0, 0, 0, 0, 1]
\end{verbatim}

\end{tcolorbox}
\caption{Description of Superply.}
\end{figure*}

\clearpage
\section{Experimental Setup}

\subsection{LLMs Configuration}
\begin{table*}[h!]
    \centering
    \renewcommand{\arraystretch}{1.2}
    \vspace{2pt}
    \scalebox{0.85}{
    \begin{tabular}{@{}cccccc@{}}
    \toprule
    \textbf{Model} & \textbf{Creator} & \textbf{Version} & \textbf{Access Time} & \textbf{License} & \textbf{Input Modalities} \\ \midrule
    \textbf{Deepseek-R1} & Deepseek & Deepseek-R1 & 2025.1 & Open-source &text \\ \midrule
    \textbf{o4-mini} & OpenAI & o4-mini-medium & 2025.4 & Proprietary & text \& image \\ \midrule
    \textbf{Gemini-2.5-pro} & Google & gemini-2.5-pro-preview-03-25 & 2025.3 & Proprietary & text \& image \\ \midrule
    \textbf{QwQ-32B} & Alibaba & QwQ-32B & 2025.3 & Open-source & text \\ \midrule
    \textbf{grok-3-mini} & xAI & grok-3-mini & 2025.2 & Proprietary & text \\ \midrule
    \textbf{Deepseek-V3} & Deepseek & Deepseek-V3 & 2024.12 & Open-source & text \\ \midrule
    \textbf{GPT-4.1} & OpenAI & gpt-4.1-2025-04-14 & 2025.4 & Proprietary & text \& image \\ \midrule
    \textbf{Qwen2.5-VL-72B} & Alibaba & Qwen2.5-VL-72B-Instruct & 2025.2 & Open-source & text \& image \\ \midrule
    \textbf{Llama-3.3-70B} & Meta & Llama-3.3-70B-Instruct & 2024.12 & Open-source & text \\ \midrule
    \textbf{Gemma-3-27B} & Google & gemma-3-27b-it & 2025.3 & Open-source & text \& image \\ \midrule
    \textbf{Phi-4-multimodal} & Microsoft & Phi-4-multimodal-instruct & 2025.3 & Open-source & text \& image \\ 
   \bottomrule
    \end{tabular}
    }
    \caption{Details of the LLMs evaluated in \ours.}
    \label{tab:judge_models}
\end{table*}

\subsection{Costomized Model Configuration}
\begin{table*}[h!]
\vspace{2pt}
\centering
\resizebox{\textwidth}{!}{
\begin{tabular}{@{}lcll@{}}
\toprule
\multirow{3}{*}{\textbf{Puzzle Name}} & & \multicolumn{2}{c}{\textbf{Baseline Strategy}} \\
\cmidrule{3-4}
&& \textbf{Easy} & \textbf{Normal} \\
\midrule
\rowcolor[HTML]{F7F7F7} 
SudoKill && Random & Greedy\\
\rowcolor[HTML]{FFFFFF} 
TidyTower && Dynamic Programming & Dynamic Programming\\
\rowcolor[HTML]{F7F7F7} 
CardNim  && Random & Dynamic Programming \\
\rowcolor[HTML]{FFFFFF} 
OptimalTouring && Simulated Annealing Algorithm & Simulated Annealing Algorithm\\
\rowcolor[HTML]{F7F7F7} 
CountMaximalCocktails && Brute-force & Brute-force \\
\rowcolor[HTML]{FFFFFF} 
MaxMaximalCocktails && Random & Brute-force \\
\rowcolor[HTML]{F7F7F7} 
ExclusivityParticles && Brute-force & Greedy \\
\rowcolor[HTML]{FFFFFF} 
ExclusivityProbes && Random & Greedy\\
\rowcolor[HTML]{F7F7F7} 
RubyRisks && Monte-Carlo Tree Search & Monte-Carlo Tree Search\\
\rowcolor[HTML]{FFFFFF} 
BeatOrBombSto. && Random & Greedy\\
\rowcolor[HTML]{F7F7F7} 
MaxTarget && Greedy & Greedy \\
\rowcolor[HTML]{FFFFFF} 
LargerTarget && Random & Greedy \\
\rowcolor[HTML]{F7F7F7} 
Superply && Random & Searching\\
\bottomrule
\end{tabular}
}
\caption{Overview of puzzle games and their basic strategies. For text-image puzzles, we apply strategies similar to those used in corresponding text-only puzzles.}
\label{tab:puzzles_table}
\end{table*}

\clearpage
\subsection{Implementation Details of Model Inference}
We use APIs to evaluate several models: Deepseek-R1, o4-mini, Gemini-2.5-pro, grok-3-mini, Deepseek-V3, GPT-4.1, and Phi-4-multimodal. For other models, we utilize Hugging Face Transformers \cite{wolf2020huggingfacestransformersstateoftheartnatural} inference on 8 $\times$ H100 and 8 $\times$ A100.

\subsection{Implementation Details of Elo Score}
\label{imple_elo}
Each player begins with an initial rating \( R = 1000 \). After a match between player A and player B, player A’s updated rating is given by $R_A' = R_A + K \cdot (S_A - E_A)$ where \( R_A \) and \( R_B \) are the current Elo ratings of players A and B, respectively.  
\( K = 32 \) is the update constant we set.  
\( S_A \in \{1, 0.5, 0\} \) is the actual result of the game (1 = win, 0.5 = draw, 0 = loss).  
\( E_A \) is the expected score for player A, computed as $E_A = \frac{1}{1 + 10^{(R_B - R_A)/400}}$.

\subsection{Operation Extraction}
In an instruction-based as well as a code-based setting, the raw output of the LLM may not be in a correct format. For each turn of an LLM, \ours allows the LLM up to five attemps  to generate a move with the correct format (based on a regular expression checker). As soon as an attempt generates a correct format, the result is sent to the state transition engine of \ours. If none of attempts generates a move having the correct format, that LLM loses the game.

\subsection{The Cost of Experiments}
\begin{table*}[!htbp]
\centering
\renewcommand{\arraystretch}{1.2} 
\setlength{\tabcolsep}{10pt}      
\large                             
\resizebox{\textwidth}{!}{
\begin{tabular}{lcccccc}
\toprule
\textbf{Model} &
\textbf{Instr.-based Cost (\$)} &
\textbf{Code-based Cost (\$)} &
\textbf{Total Cost (\$)} &
\textbf{Instr.-based GPU Hrs} &
\textbf{Code-based GPU Hrs} &
\textbf{Total GPU Hrs} \\
\midrule
GPT-4.1            & 162.83 &  7.59 & 170.42 & -- & -- & -- \\
o4-mini            &  90.03 &  3.27 &  93.30 & -- & -- & -- \\
Gemini-2.5-pro     & 556.86 & 33.03 & 589.89 & -- & -- & -- \\
grok-3-mini        &  20.16 &  5.21 &  25.37 & -- & -- & -- \\
Deepseek-R1        &  37.41 & 14.78 &  52.19 & -- & -- & -- \\
Deepseek-V3        &  27.99 &  1.22 &  29.21 & -- & -- & -- \\
Phi-4-multimodal   &   8.82 &  0.36 &   9.18 & -- & -- & -- \\
QwQ-32B            & -- & -- & -- & 50.82 & 6.76 & 57.58 \\
Llama-3.3-70B      & -- & -- & -- & 48.17 & 2.02 & 50.19 \\
Gemma-3-27B        & -- & -- & -- & 10.17 & 3.89 & 14.06 \\
Qwen2.5-VL-72B     & -- & -- & -- & 56.30 & 1.96 & 58.26 \\
\midrule
\textbf{Total}     & 903.10 & 65.46 & 968.56 & 165.46 & 14.63 & 180.09 \\
\bottomrule
\end{tabular}%
}
\caption{Experiment costs estimated across both instruction-based and code-based settings using two approaches: API-based models are priced in USD, and GPU-based models are quantified in NVIDIA H100 GPU hours.}
\label{tab:model_costs}
\end{table*}

\subsection{Code-based Prompt Template}
\label{code_template}
\begin{figure*}[h!]
\begin{tcolorbox}[colback=white, colframe=blue!50!black,  title=Code Template Prompt, fontupper=\footnotesize, fonttitle=\footnotesize]
You are about to play a game called \{puzzle name\} against an opponent. \\

\{puzzle rules\} \\

\{code input description and format\} \\

\{code output description and format\} \\

\{code template\} \\

\{required LLM output format\}

\end{tcolorbox}
\caption{The code template prompt in code-based setting.}
\end{figure*}

\clearpage
\section{Experiment Results}
\subsection{Elo Score Results}
\label{more_results}

\begin{table}[h]
\centering
\small
\resizebox{0.9\linewidth}{!}{
\begin{tabular}{lcccccc}
\toprule
\multirow{2}{*}{\textbf{Model}} &
\multicolumn{2}{c}{\textbf{Single-Player Det.}} &
\multicolumn{2}{c}{\textbf{Two-Player Det.}} &
\multirow{2}{*}{\textbf{Score}} \\
\cmidrule(lr){2-3}  \cmidrule(lr){4-5}
 & Easy & Normal & Easy & Normal \\
\midrule
Custom & 1250.9 \tiny{$\pm$504.5} & 1229.9 \tiny{$\pm$529.1} & 1060.8 \tiny{$\pm$309.8} & 1081.3 \tiny{$\pm$347.2} & 1134.6 \tiny{$\pm$126.4} \\
\midrule
Deepseek-R1 & \cellcolor{red!35}1146.6 \tiny{$\pm$445.8} & \cellcolor{red!20}1084.3 \tiny{$\pm$379.5} & \cellcolor{red!20}1163.9 \tiny{$\pm$122.0} & \cellcolor{red!35}1185.1 \tiny{$\pm$126.2} & \cellcolor{red!35}1152.4 \tiny{$\pm$63.2} \\
o4-mini & \cellcolor{red!20}1094.9 \tiny{$\pm$549.8} & \cellcolor{red!35}1120.0 \tiny{$\pm$649.5} & \cellcolor{red!35}1165.7 \tiny{$\pm$96.0} & \cellcolor{red!20}1156.4 \tiny{$\pm$100.3} & \cellcolor{red!20}1140.9 \tiny{$\pm$74.9} \\
Gemini-2.5-pro & \cellcolor{red!5}1082.1 \tiny{$\pm$494.9} & \cellcolor{red!5}1046.3 \tiny{$\pm$241.2} & 1145.0 \tiny{$\pm$64.3} & 1117.8 \tiny{$\pm$136.7} & \cellcolor{red!5}1106.2 \tiny{$\pm$58.1} \\
QwQ-32B & 1074.1 \tiny{$\pm$268.5} & 988.2 \tiny{$\pm$269.8} & \cellcolor{red!5}1160.0 \tiny{$\pm$111.5} & \cellcolor{red!5}1122.9 \tiny{$\pm$84.4} & 1100.1 \tiny{$\pm$54.7} \\
grok-3-mini & 882.4 \tiny{$\pm$366.0} & 911.1 \tiny{$\pm$254.8} & 1145.9 \tiny{$\pm$190.1} & 1120.6 \tiny{$\pm$225.1} & 1044.6 \tiny{$\pm$97.5} \\
Deepseek-V3 & 980.2 \tiny{$\pm$60.4} & 950.7 \tiny{$\pm$89.1} & 989.6 \tiny{$\pm$163.4} & 967.7 \tiny{$\pm$87.9} & 973.7 \tiny{$\pm$42.7} \\
GPT-4.1 & 1041.8 \tiny{$\pm$321.8} & 1042.5 \tiny{$\pm$321.4} & 1024.8 \tiny{$\pm$137.2} & 1044.7 \tiny{$\pm$111.1} & 1037.5 \tiny{$\pm$53.1} \\
Qwen-2.5-VL-72B & 905.2 \tiny{$\pm$349.7} & 950.7 \tiny{$\pm$282.0} & 762.6 \tiny{$\pm$117.0} & 817.2 \tiny{$\pm$194.8} & 841.7 \tiny{$\pm$73.0} \\
Llama-3.3-70B & 863.3 \tiny{$\pm$381.6} & 891.9 \tiny{$\pm$266.0} & 826.9 \tiny{$\pm$108.3} & 792.1 \tiny{$\pm$99.6} & 835.0 \tiny{$\pm$52.9} \\
Gemma-3-27B & 858.8 \tiny{$\pm$397.7} & 918.3 \tiny{$\pm$217.5} & 797.5 \tiny{$\pm$114.3} & 797.7 \tiny{$\pm$103.2} & 831.7 \tiny{$\pm$55.8} \\
Phi-4-multimodal & 819.8 \tiny{$\pm$294.0} & 866.0 \tiny{$\pm$322.9} & 757.2 \tiny{$\pm$144.9} & 796.5 \tiny{$\pm$97.6} & 801.6 \tiny{$\pm$55.8} \\
\bottomrule
\end{tabular}
}
\caption{Comparison of Elo scores (mean $\pm$ 95\% CI) across various models on single-player and two-player deterministic puzzles, categorized by difficulty. 
}
\label{tab:det_puzzle_scores_updated}
\end{table}
\begin{table}[ht]
\centering
\small
\begin{tabular}{lcccccc}
\toprule
\multirow{2}{*}{\textbf{Model}} &
\multicolumn{2}{c}{\textbf{TidyTower}} &
\multicolumn{2}{c}{\textbf{OptimalTouring}} &
\multicolumn{2}{c}{\textbf{CountMaximalCocktails}} \\
\cmidrule(lr){2-3}  \cmidrule(lr){4-5} \cmidrule(lr){6-7}
 & Easy & Normal & Easy & Normal & Easy & Normal\\
\midrule
{Custom} & {1481.3} & {1464.0} & {1098.1} & {1047.6} & {1173.2} & {1178.2} \\
\midrule
{o4-mini} & 956.1 & 952.8 & \cellcolor{red!35}{1350.1} & \cellcolor{red!35}{1421.3} & 978.4 & {985.8} \\
{Deepseek-r1} & 956.2 & 952.4 & \cellcolor{red!20}{1312.6} & \cellcolor{red!20}{1251.6} & \cellcolor{red!35}{1170.9} & \cellcolor{red!35}{1048.8} \\
{Deepseek-V3} & 956.2 & 952.2 & {1004.8} & 914.1 & 979.7 & 985.8 \\
{GPT-4.1} & 956.1 & 953.0 & 1190.8 & \cellcolor{red!5}{1190.9} & 978.4 & 983.8 \\
{grok-3-mini} & 956.2 & 952.5 & 712.7 & 794.3 & 978.2 & 986.5 \\
{Gemini-2.5-pro} & 956.1 & 952.7 & \cellcolor{red!5}{1311.8} & 1146.5 & 978.3 & \cellcolor{red!20}{1039.8} \\
{QwQ-32B} & 956.3 & 952.0 & 1168.7 & 1110.3 & \cellcolor{red!20}{1097.2} & 902.4 \\
{Qwen-2.5-VL-72B} & \cellcolor{red!35}{956.4} & \cellcolor{red!35}{1012.7} & 746.0 & 819.8 & \cellcolor{red!5}{1013.2} & \cellcolor{red!5}{1019.8} \\
{Llama-3.3-70B} & 956.3 & 951.9 & 686.0 & 768.3 & 947.5 & 955.5 \\
{Gemma-3-27B} & \cellcolor{red!35}{956.4} & 951.8 & 674.1 & 819.0 & 946.1 & 984.2 \\
{Phi-4-multimodal} & 956.3 & 952.1 & 744.4 & 716.5 & 758.9 & 929.4 \\
\bottomrule
\end{tabular}
\caption{Performance comparison in Elo ratings of large language models on three single-player deterministic puzzles, categorized by difficulty. 
}
\label{tab:selected_puzzles}
\end{table}

\begin{table}[h!]
\centering
\resizebox{\textwidth}{!}{
\begin{tabular}{lcccccccccc}
\toprule
\multirow{2}{*}{\textbf{Model}} &
\multicolumn{2}{c}{\textbf{CardNim}} &
\multicolumn{2}{c}{\textbf{SudoKill}} &
\multicolumn{2}{c}{\textbf{MaxMaximalCocktails}} &
\multicolumn{2}{c}{\textbf{ExclusivityParticles}} &
\multicolumn{2}{c}{\textbf{Superply}} \\
\cmidrule(lr){2-3} \cmidrule(lr){4-5} \cmidrule(lr){6-7} \cmidrule(lr){8-9} \cmidrule(lr){10-11}
& Easy & Normal & Easy & Normal & Easy & Normal & Easy & Normal & Easy & Normal  \\
\midrule
{Custom} & 688.9 & 697.6 & 1155.1 & 1355.5 & 1274.6 & 1106.5 & 1257.2 & 1331.1 & 928.2 & 915.8 \\
\midrule
{o4-mini} & {1206.8} & {1153.9} & \cellcolor{red!5}{1149.5} & \cellcolor{red!5}{1048.4} & \cellcolor{red!5}{1039.1} & \cellcolor{red!35}{1124.0} & \cellcolor{red!35}{1196.9} & \cellcolor{red!35}{1188.1} & \cellcolor{red!20}{1236.2} & \cellcolor{red!35}{1267.8} \\
{Deepseek-r1} & \cellcolor{red!5}{1245.8} & \cellcolor{red!20}{1265.2} & \cellcolor{red!5}{1050.3} & \cellcolor{red!35}{1266.1} & \cellcolor{red!35}{1110.9} & \cellcolor{red!5}{1025.3} & \cellcolor{red!20}{1126.8} & \cellcolor{red!20}{1146.8} & \cellcolor{red!35}{1285.6} & \cellcolor{red!20}{1222.4} \\
{Deepseek-V3} & 1042.0 & 1002.4 & \cellcolor{red!20}{1198.3} & 964.0 & 898.5 & 1012.8 & 913.8 & 846.3 & 895.4 & 1013.1 \\
{GPT-4.1} & 984.1 & 1084.2 & 1071.9 & \cellcolor{red!5}{1033.7} & 999.3 & 990.0 & 885.7 & 941.5 & 1182.9 & \cellcolor{red!5}{1173.9} \\
{grok-3-mini} & \cellcolor{red!35}{1340.8} & \cellcolor{red!35}{1313.8} & \cellcolor{red!35}{1216.0} & \cellcolor{red!20}{1260.2} & 935.0 & 864.6 & 1069.1 & 1026.6 & 1168.9 & 1137.7 \\
{Gemini-2.5-pro} & 1201.7 & 1221.1 & 1160.3 & 1090.6 & \cellcolor{red!20}{1063.4} & \cellcolor{red!20}{1109.0} & \cellcolor{red!5}{1166.5} & \cellcolor{red!35}{1216.1} & 1133.0 & 952.4 \\
{QwQ-32B} & \cellcolor{red!20}{1252.2} & \cellcolor{red!5}{1190.7} & 1132.3 & 1116.5 & \cellcolor{red!5}{1021.2} & 1016.3 & \cellcolor{red!5}{1175.4} & \cellcolor{red!5}{1118.5} & \cellcolor{red!5}{1219.1} & \cellcolor{red!5}{1172.4} \\
{Qwen-2.5-VL-72B} & 679.4 & 789.9 & 705.5 & 659.5 & 873.3 & \cellcolor{red!5}{1081.2} & 857.0 & 782.6 & 698.0 & 772.9 \\
{Llama-3.3-70B} & 839.8 & 847.7 & 789.3 & 653.9 & 971.4 & 847.8 & 786.8 & 811.6 & 747.1 & 799.3 \\
{Gemma-3-27B} & 890.8 & 752.7 & 662.3 & 771.3 & 875.7 & 936.1 & 771.4 & 806.1 & 787.4 & 722.3 \\
{Phi-4-multimodal} & 627.6 & 680.9 & 709.1 & 780.3 & 937.6 & 886.4 & 793.2 & 784.8 & 718.3 & 850.1 \\
\bottomrule
\end{tabular}
}
\caption{Performance comparison in Elo ratings of large language models on five two-player deterministic puzzles. Scores are reported in Elo ratings for both Easy and Normal difficulty settings.
}
\label{tab:additional_puzzles}
\end{table}

\clearpage
\subsection{Results of Different Prompting Strategies}
\begin{table}[ht]
    \centering
    \small
    \addtolength{\tabcolsep}{-0.3em}
    \resizebox{0.45\linewidth}{!}{
    \begin{tabular}{lcccccccccccc}
    \toprule
    \multirow{2}{*}{\textbf{Model}} &
    \multicolumn{2}{c}{\textbf{TidyTower}} &
    \multicolumn{2}{c}{\textbf{SudoKill}} \\
    \cmidrule(lr){2-3} \cmidrule(lr){4-5}
    & \textbf{Easy} & \textbf{Normal} & \textbf{Easy} & \textbf{Normal} \\ 
    \midrule
    GPT-4.1 & 0.00 & 0.00 & 0.30 & 0.10 \\
    \noalign{\vskip 0.5ex}  \cdashline{1-5} \noalign{\vskip 0.5ex}
    \quad $w/$ 1-shot  & 0.00 & 0.00 & 0.10 & 0.10\\
    \noalign{\vskip 0.5ex}   \noalign{\vskip 0.5ex}
    \quad $w/$ ToT  & 0.60 & 0.80 & 0.30 & 0.10\\
    \noalign{\vskip 0.5ex}   \noalign{\vskip 0.5ex}
    \quad $w/o$ history & 1.00 & 0.90 & 0.20 & 0.00\\
    \noalign{\vskip 0.5ex}  \noalign{\vskip 0.5ex}
    \quad $w/$ legal candidates & 0.60 & 0.00 & 0.60 & 0.70 \\
    \midrule
    o4-mini & 0.00 & 0.00 & 0.40 & 0.20 \\
    \noalign{\vskip 0.5ex}  \cdashline{1-5} \noalign{\vskip 0.5ex}
    \quad $w/$ 1-shot & 0.00 & 0.00 & 0.50 & 0.20  \\
    \noalign{\vskip 0.5ex}\noalign{\vskip 0.5ex}
    \quad $w/$ ToT & 0.80 & 0.70 & 0.40 & 0.40 \\
    \noalign{\vskip 0.5ex}\noalign{\vskip 0.5ex}
    \quad $w/o$ history & 1.00 & 1.00 & 0.40 & 0.30 \\
    \noalign{\vskip 0.5ex} \noalign{\vskip 0.5ex}
    \quad $w/$ legal candidates & 1.00 & 0.60 & 0.50 & 0.40 \\
    \bottomrule
    \end{tabular}
    }
        \caption{Performance of GPT-4.1 and o4-mini on \textsc{TidyTower} and \textsc{SudoKill} puzzles under different prompting strategies.}
    \label{tab:ablation}
\end{table}

\subsection{Breakdown Instruction-based Results of Puzzles}
\begin{table}[!h]
\small
\centering
\begin{tabular}{lcccccc}
\toprule
\multirow{2}{*}{\textbf{Model}} &
\multicolumn{2}{c}{\textbf{TidyTower}} &
\multicolumn{2}{c}{\textbf{OptimalTouring}} &
\multicolumn{2}{c}{\textbf{CountMaximalCocktails}} \\
\cmidrule(lr){2-3}  \cmidrule(lr){4-5} \cmidrule(lr){6-7}
 & Easy & Normal & Easy & Normal & Easy & Normal\\
\midrule
{Custom} & 1.00 & 1.00 & 0.67 & 0.49 & 1.00 & 1.00 \\
\midrule
{Deepseek-R1} & 0.00 & 0.00 & \cellcolor{red!20}{0.91} & \cellcolor{red!5}{0.75} & \cellcolor{red!35}{1.00} & \cellcolor{red!35}{0.70}\\
{o4-mini} & 0.00 & 0.00 & \cellcolor{red!35}{0.93} & \cellcolor{red!35}{0.92} & 0.40 & \cellcolor{red!5}{0.40} \\
{Gemini-2.5-pro} & 0.00 & 0.00 & \cellcolor{red!20}{0.91} & \cellcolor{red!20}{0.82} & 0.40 & \cellcolor{red!20}{0.50} \\
{QwQ-32B} & 0.00 & 0.00 & {0.81} & 0.46 & \cellcolor{red!20}{0.80} & 0.30\\
{grok-3-mini} & 0.00 & 0.00 & 0.10 & 0.27 & 0.40 & \cellcolor{red!5}{0.40}\\
{Deepseek-V3} & 0.00 & 0.00 & 0.62 & 0.42 & 0.40 & 0.30 \\
{GPT-4.1} & 0.00 & 0.00 & 0.79 & \cellcolor{red!5}{0.75} & 0.40 & 0.30\\
{Qwen-2.5-VL-72B} & 0.00 & \cellcolor{red!35}{0.20} & 0.21 & 0.11 & \cellcolor{red!5}{0.50} & \cellcolor{red!5}{0.40}\\
{Llama-3.3-70B} & 0.00 & 0.00 & 0.14 & 0.07 & 0.30 & 0.30\\
{Gemma-3-27B} & 0.00 & 0.00 & 0.09 & 0.06  & 0.30 & 0.30\\
{Phi-4-multimodal} & 0.00 & 0.00 & 0.14 & 0.00 & 0.00 & 0.10\\
\bottomrule
\end{tabular}
\caption{Instruction-based normalized scores of models on single-player deterministic puzzles, separated by difficulty.
}
\label{tab:single_det_puzzle_scores}
\end{table}
\begin{table}[!h]
\centering
\small
\resizebox{1.0\linewidth}{!}{
\begin{tabular}{lcccccccccc}
\toprule
\multirow{2}{*}{\textbf{Model}} &
\multicolumn{2}{c}{\textbf{CardNim}} &
\multicolumn{2}{c}{\textbf{SudoKill}} &
\multicolumn{2}{c}{\textbf{MaxMaximalCocktails}} &
\multicolumn{2}{c}{\textbf{ExclusivityParticles}} &
\multicolumn{2}{c}{\textbf{Superply}} \\
\cmidrule(lr){2-3} \cmidrule(lr){4-5} \cmidrule(lr){6-7} \cmidrule(lr){8-9} \cmidrule(lr){10-11}
& Easy & Normal & Easy & Normal & Easy & Normal & Easy & Normal & Easy & Normal  \\
\midrule
{Custom}  & 0.16 & 0.25 & 0.75 & 0.86 & 0.83 & 0.64 & 0.80 & 0.86 & 0.43 & 0.40 \\
\midrule
{Deepseek-R1}  & \cellcolor{red!5}{0.77} & \cellcolor{red!5}{0.76} & 0.64 & \cellcolor{red!20}{0.74} & \cellcolor{red!20}{0.57} & 0.54 & 0.56 & 0.54 & 0.74 & \cellcolor{red!5}{0.73} \\
{o4-mini} & 0.73 & 0.63 & 0.62 & 0.59 & \cellcolor{red!5}{0.51} & \cellcolor{red!35}{0.65} & \cellcolor{red!5}{0.68} & \cellcolor{red!20}{0.70} & \cellcolor{red!35}{0.83} & \cellcolor{red!35}{0.83} \\
{Gemini-2.5-pro} & \cellcolor{red!20}{0.79} & \cellcolor{red!20}{0.79} & \cellcolor{red!35}{0.75} & \cellcolor{red!5}{0.64} & 0.50 & \cellcolor{red!5}{0.60} & \cellcolor{red!35}{0.70} & \cellcolor{red!35}{0.75} & 0.66 & 0.55 \\
{QwQ-32B}  & 0.76 & 0.73 & 0.65 & 0.61 & \cellcolor{red!35}{0.59} & 0.57 & \cellcolor{red!20}{0.69} & \cellcolor{red!20}{0.70} & \cellcolor{red!20}{0.78} & \cellcolor{red!20}{0.78} \\
{grok-3-mini}  & \cellcolor{red!35}{0.83} & \cellcolor{red!35}{0.83} & \cellcolor{red!20}{0.70} & \cellcolor{red!35}{0.79} & 0.41 & 0.42 & 0.65 & \cellcolor{red!5}{0.60} & \cellcolor{red!5}{0.76} & 0.72 \\
{Deepseek-V3} & 0.53 & 0.54 & \cellcolor{red!5}{0.69} & 0.57 & 0.45 & 0.52 & 0.46 & 0.30 & 0.45 & 0.56 \\
{GPT-4.1}  & 0.47 & 0.46 & 0.45 & 0.40 & 0.47 & 0.44 & 0.25 & 0.38 & 0.53 & 0.56 \\
{Qwen-2.5-VL-72B} & 0.13 & 0.26 & 0.18 & 0.15 & 0.44 & \cellcolor{red!20}{0.61} & 0.34 & 0.23 & 0.12 & 0.16 \\
{Llama-3.3-70B}  & 0.34 & 0.39 & 0.23 & 0.17 & 0.46 & 0.40 & 0.31 & 0.33 & 0.22 & 0.26 \\
{Gemma-3-27B} & 0.27 & 0.19 & 0.13 & 0.24 & 0.36 & 0.38 & 0.15 & 0.25 & 0.23 & 0.17 \\
{Phi-4-multimodal}  & 0.16 & 0.13 & 0.10 & 0.14 & 0.33 & 0.18 & 0.18 & 0.18 & 0.09 & 0.24 \\
\bottomrule
\end{tabular}
}
\caption{Instruction-based normalized scores of models on two-player deterministic puzzles, separated by difficulty.}
\label{tab:two_det_puzzle_scores}
\end{table}

\clearpage
\subsection{Instruction-based vs. Code-based}
\begin{table}[ht]
\centering
\small
\begin{tabular}{lcc}
\toprule
\textbf{Model} & \textbf{Easy} & \textbf{Normal} \\
\midrule
Deepseek-R1 & 0.56 \textcolor{blue}{(-0.25)} & 0.50 \textcolor{blue}{(-0.32)} \\
o4-mini & 0.62 \textcolor{blue}{(-0.21)} & 0.54 \textcolor{blue}{(-0.19)} \\
Gemini-2.5-pro & 0.64 \textcolor{blue}{(-0.24)} & 0.58 \textcolor{blue}{(-0.23)} \\
QwQ-32B  & 0.48 \textcolor{blue}{(-0.26)} & 0.33 \textcolor{blue}{(-0.25)} \\
grok-3-mini & 0.64 \textcolor{blue}{(-0.30)} & 0.56 \textcolor{blue}{(-0.28)} \\
Deepseek-V3 & 0.40 \textcolor{blue}{(-0.27)} & 0.32 \textcolor{blue}{(-0.26)} \\
GPT-4.1 & 0.46 \textcolor{blue}{(-0.25)} & 0.36 \textcolor{blue}{(-0.27)} \\ 
Qwen-2.5-VL-72B & 0.30 \textcolor{blue}{(-0.21)} & 0.24 \textcolor{blue}{(-0.19)} \\
Llama-3.3-70B  & 0.32 \textcolor{blue}{(-0.16)} & 0.07 \textcolor{blue}{(-0.07)} \\
Gemma-3-27B  & 0.06 \textcolor{blue}{(-0.05)} & 0.00 \textcolor{blue}{(-0.00)} \\
Phi-4-multimodal  & 0.18 \textcolor{blue}{(-0.18)} & 0.20 \textcolor{blue}{(-0.20)} \\
\bottomrule
\end{tabular}
\caption{Win rates of foundation models against the custom strategy in the instruction-based setting on two-player deterministic puzzles. The blue values in parentheses indicate the win rate difference (instruction-based minus code-based), highlighting performance drops in the code-based setting.}
\label{tab:code_instruction_comparison}
\end{table}

\subsection{Play Statistics}
\begin{table*}[h!]
\centering
\renewcommand{\arraystretch}{1.35} 
\setlength{\tabcolsep}{8pt}       
\large                              

\resizebox{\linewidth}{!}{ 
\begin{tabular}{lcccccccc}
\toprule
\multirow{2}{*}{\textbf{Name}} &
\multicolumn{3}{c}{\textbf{Total Play}} &
\multicolumn{3}{c}{\textbf{Legal Play}} &
\multirow{2}{*}{\textbf{Legal Play Percentage}} \\
\cmidrule(lr){2-4} \cmidrule(lr){5-7}
& \#Turns & \#Tokens (R) & \#Tokens (NR) &
\#Turns & \#Tokens (R) & \#Tokens (NR) &  \\

\midrule
SudoKill & 3.26 $\pm$ 0.11 & 6281.64 $\pm$ 73.58 & 916.10 $\pm$ 24.00 & 7.40 $\pm$ 0.39 & 5362.98 $\pm$ 118.59 & 812.60 $\pm$ 38.14 & 0.12 \\
TidyTower & 1.00 $\pm$ 0.00 & 7636.68 $\pm$ 334.95 & 1273.59 $\pm$ 76.42 & 1.00 $\pm$ 0.00 & 7636.68 $\pm$ 334.95 & 1282.83 $\pm$ 77.25 & 0.99 \\
CardNim & 1.79 $\pm$ 0.02 & 5690.17 $\pm$ 143.70 & 492.85 $\pm$ 16.51 & 1.85 $\pm$ 0.02 & 5383.76 $\pm$ 156.37 & 444.68 $\pm$ 12.80 & 0.77 \\
OptimalTouring & 1.00 $\pm$ 0.00 & 17370.60 $\pm$ 615.55 & 1448.76 $\pm$ 70.53 & 1.00 $\pm$ 0.00 & 16375.78 $\pm$ 706.73 & 1563.17 $\pm$ 103.28 & 0.62 \\
CountMaximalCocktails & 1.00 $\pm$ 0.00 & 5217.48 $\pm$ 441.32 & 1228.92 $\pm$ 94.87 & 1.00 $\pm$ 0.00 & 5268.49 $\pm$ 459.68 & 1256.98 $\pm$ 96.57 & 0.96 \\
MaxMaximalCocktails & 1.56 $\pm$ 0.02 & 7286.40 $\pm$ 150.85 & 717.81 $\pm$ 21.12 & 1.51 $\pm$ 0.02 & 7870.05 $\pm$ 167.68 & 736.76 $\pm$ 23.98 & 0.78 \\
ExclusivityParticles & 7.09 $\pm$ 0.49 & 2823.25 $\pm$ 39.29 & 445.75 $\pm$ 6.75 & -- & -- & -- & 0.00 \\
Superply & 10.13 $\pm$ 0.20 & 2519.22 $\pm$ 26.78 & 344.56 $\pm$ 4.17 & 11.39 $\pm$ 0.20 & 2514.00 $\pm$ 27.10 & 333.38 $\pm$ 3.40 & 0.84 \\
\bottomrule
\end{tabular}
}
\caption{Statistics of play and legal play across puzzles in the instruction-based setting. (R) denotes reasoning models, and (NR) denotes non-reasoning models. A legal play refers to a game trajectory that ends with a legal termination status; for a detailed definition of termination status, please refer to \S~\ref{status}. \#Turns indicates the number of turns per player. In two-player puzzles, the total number of rounds is the sum of turns across both players.}
\label{tab:legal_play_stats}
\end{table*}

\clearpage
\subsection{Win Probability Matrix}
\begin{figure}[!h]
    \centering
    \includegraphics[width=1.00\textwidth]{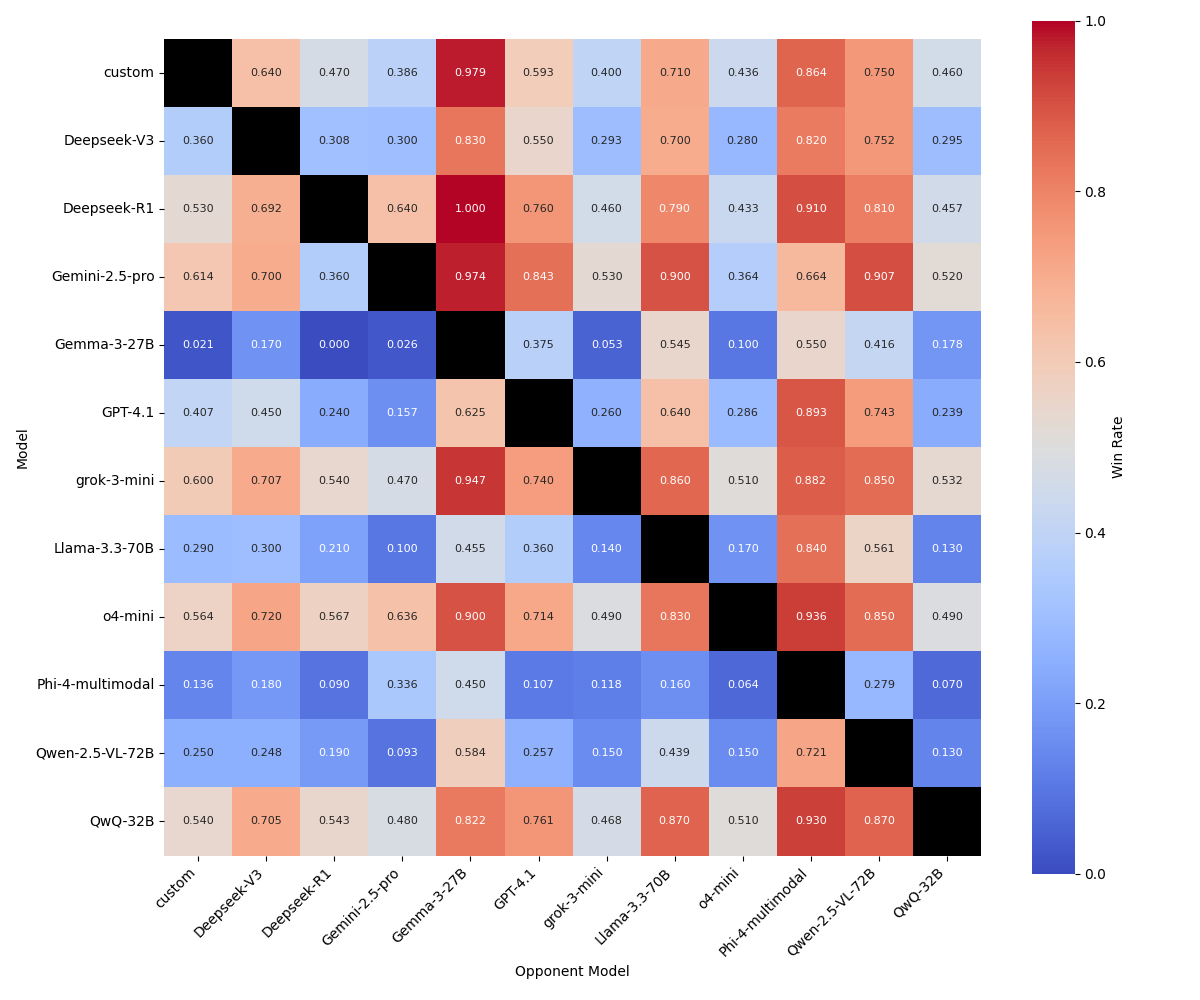}
    \caption{Win rate heatmap for 12 methods in two-player puzzles under the \textbf{instruction-based} setting. Each number represents the win rate of the row entry over the column entry, normalized to ignore ties. For example, for game instances that don't end in a tie, GPT-4.1 beats o4-mini 28.6\% of the time and o4-mini beats GPT-4.1 71.4\% of the time. }
    \label{fig:win_rate}
\end{figure}

\begin{figure}[!h]
    \centering
    \includegraphics[width=1.0\textwidth]{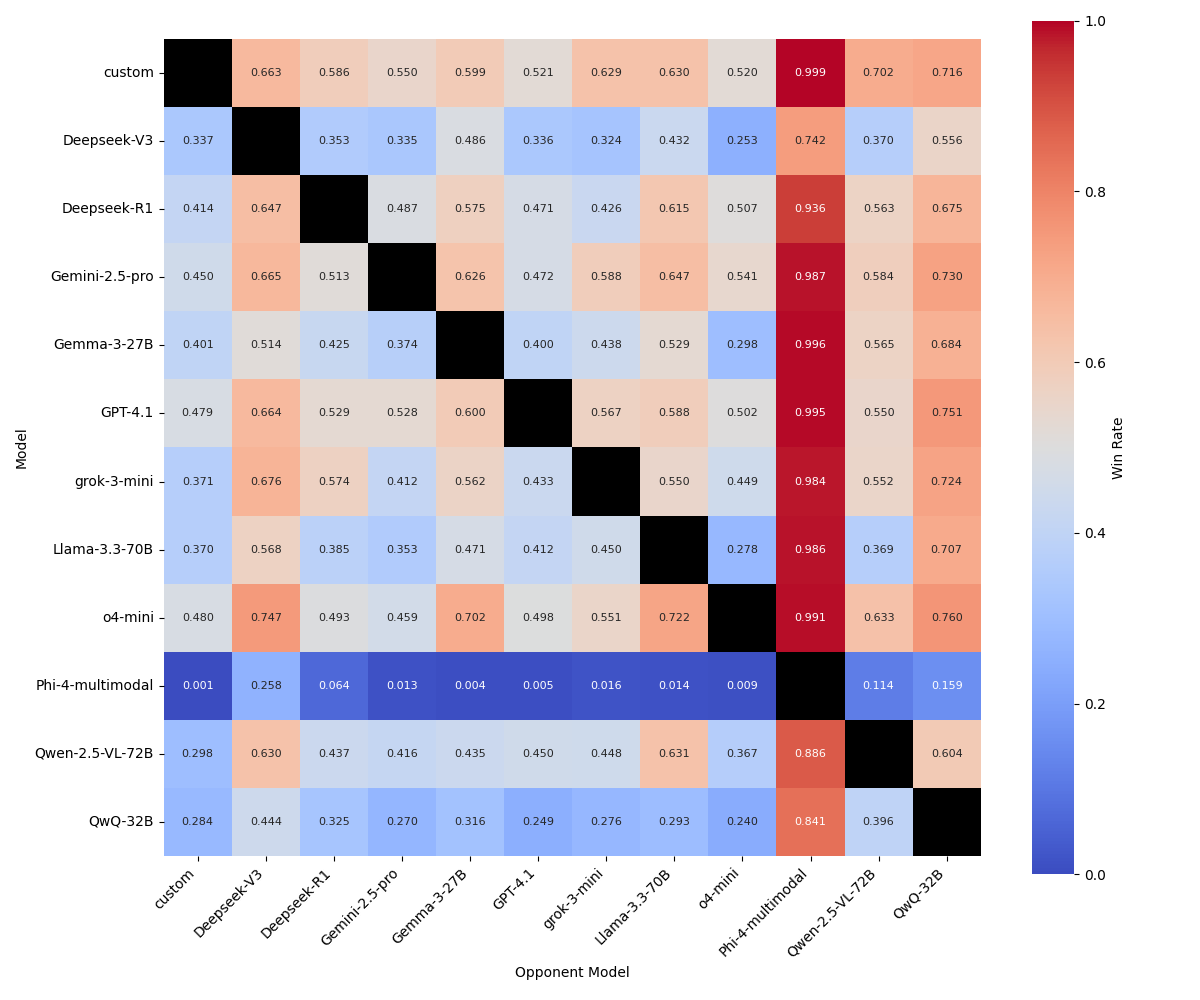}
    \caption{
    Win rate heatmap for 12 methods in two-player puzzles under the \textbf{code-based} setting. Each number represents the win rate of the row entry over the column entry, normalized to ignore ties. For example, for game instances that don't end in a tie, GPT-4.1 beats o4-mini 50.2\% of the time and o4-mini beats GPT-4.1 49.8\% of the time. }
    \label{fig:win_rate_code}
\end{figure}

\end{document}